%% file: main.tex
\documentclass[12pt]{l4dc2023}
\DeclareMathAlphabet{\mathpzc}{OT1}{pzc}{m}{it}

\title[CLAS: Central Latent Action Spaces]{CLAS: Coordinating Multi-Robot Manipulation\\ with Central Latent Action Spaces}
\usepackage{times}
\usepackage{mlmacros}       
\variables{x,a,u,v,o}
\variables[mean,std]{\mu,\sigma}
\varmacros{E}{\mathcal{E}}

\probdists{p,q}
\probdists[policy]{\pi}
\MkProbDist{P}{\mathbb{P}}

\usepackage{amsfonts}

\probdists{p,q}
\probdists[policy]{\pi}
\MkProbDist{P}{\mathbb{P}}

\usepackage{fancyvrb}

\usepackage[utf8]{inputenc} 

\usepackage{todonotes}
\setuptodonotes{inline}
\usepackage{changes}

\usepackage{enumitem}
\usepackage{tikz}
\usepackage{svg}
\usetikzlibrary{fit,arrows,shapes,positioning,decorations.pathmorphing,decorations.pathreplacing,mindmap,backgrounds}
\usetikzlibrary{matrix}
\usetikzlibrary{bayesnet}
\usetikzlibrary{calc}
\usepackage{pgfplots}

\usepackage{caption}
\author{%
 \Name{Elie Aljalbout} \Email{elie.aljalbout@argmax.ai}
 \AND
 \Name{Maximilian Karl} \Email{karlma@argmax.ai}
 \AND
 \Name{Patrick van der Smagt} \\
 \addr Machine Learning Research Lab, Volkswagen Group, Munich, Germany%
}

\begin{document}
\maketitle

\begin{abstract}%
     Multi-robot manipulation tasks involve various control entities that can be separated into dynamically independent parts.
    A typical example of such real-world tasks is dual-arm manipulation.
    Learning to naively solve such tasks with reinforcement learning is often unfeasible due to the sample complexity and exploration requirements growing with the dimensionality of the action and state spaces.
    Instead, we would like to handle such environments as multi-agent systems and have several agents control parts of the whole.
    However, decentralizing the generation of actions requires coordination across agents through a channel limited to information central to the task.
    This paper proposes an approach to coordinating multi-robot manipulation through learned latent action spaces that are shared across different agents. 
    We validate our method in simulated multi-robot manipulation tasks and demonstrate improvement over previous baselines in terms of sample efficiency and learning performance.
\end{abstract}


\begin{figure}[htb]
    \centering
    \vspace{-0.3em}
    \includegraphics[width=0.75\textwidth]{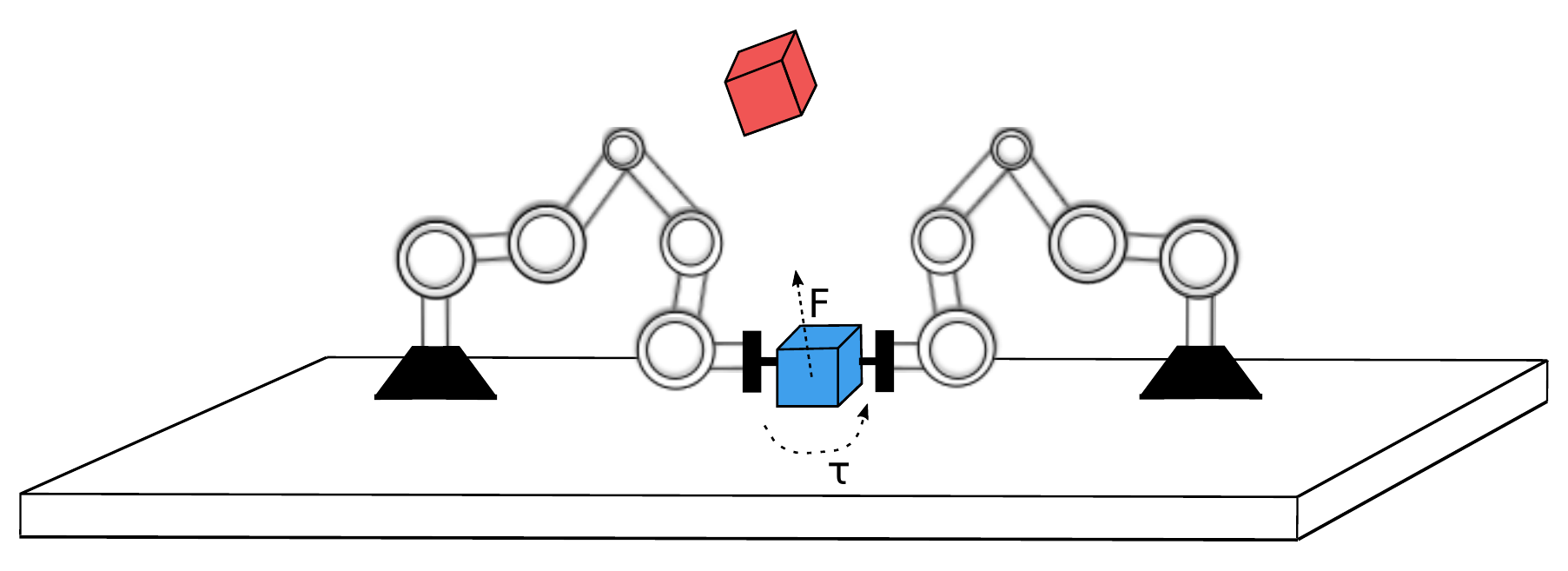}
    \vspace{-0.3em}
    \caption{Two robot arms cooperating on an object lifting task. The red cube indicates the target pose. Traditionally, two agents would control the separate robot arms in a control space of the robot such as joint torque control. We explore the option of learning latent central actions spaces which are robot-agnostic and central to the task. In our example, a possible action space would correspond to the force $F$ and torque $\tau$ acting on the center of mass of the cube.}
    \vspace{-1em}
    \label{fig:pullfig}
\end{figure}

\input{L4DC/sections/introduction.tex}
\input{L4DC/sections/related}
\input{L4DC/sections/method}

\input{L4DC/sections/experiments}

\input{L4DC/sections/conclusion}

\bibliography{references}

\appendix

\input{L4DC/sections/appendix}

\end{document}

%% file: L4DC/sections/introduction.tex
\section{Introduction}
\label{sec:introd}
\thispagestyle{plain}

Most recent successes of reinforcement learning (RL) methods have been in single-agent environments. 
Applications include games~\citep{mnih2013playing, silver2016mastering}, robotics~\citep{kober2013reinforcement}, and autonomous driving~\citep{kiran2021deep}. 
However, many control problems can naturally be distributed to multiple agents. 
For instance, in robotics, tasks involving multiple robots can be framed as multi-agent systems. 
In this work, we consider cooperative multi-agent systems. 
Such environments can also be framed as single-agent RL problems, with policies receiving full observations and outputting a single action responsible for actuating all the different entities. 
However, such an approach would suffer from great sample complexity due to the difficulty of exploration and fitting a policy under high-dimensional action, state, and observation spaces.
Instead, in a multi-agent reinforcement learning (MARL) approach, each agent is responsible for actuating a sub-part of the environment and could have access to either all observations or a subset. This simplifies the exploration and sample requirement for each individual agent.

However, multi-agent methods suffer from the lack of information present to each agent, which results in multiple problems~\citep{grana2011cooperative, canese2021multi}. 
Hence, in the literature, multiple solutions have been proposed to approach these problems. 
Most of these methods attempt to establish either an explicit or implicit communication channel between multiple agents, allowing for sharing a certain amount of information that is assumed important to the task. 
The latter refers to obtaining information about other agents through learned approximate models, not direct communication. 
Examples of such methods include opponent modeling~\citep{raileanu2018modeling,liu2020multi,yu2021model} and latent intention/goal estimation~\citep{xie2020learning,wang2020model}.

Another challenge for multi-agent systems is the decentralized action generation. 
This aspect can be ignored for the classical application domains examined by previous MARL research, such as games and particle environments. 
However, it becomes critical when dealing with physical tasks, such as dual-arm manipulation, where decoupled actions could lead to instabilities and even damage the robots.
Hence, in this work, we focus on this problem.

We postulate that there exists an agent-agnostic latent action space that can alternatively be used for solving certain families of cooperative multi-robot manipulation tasks. 
To illustrate the concept, we take the example of multiple robot manipulators lifting a single object. 
In the standard multi-agent approach to solving this task, each agent would be responsible for actuating one robot. 
Each agent's action space would correspond to some control commands suitable for the robot, \eg joint torques, velocities, Cartesian poses. 
The goal of this task is to move the object; hence a simpler and more intuitive action space would ideally be expressed with respect to the object and not the robots.
For instance, such an action could represent a wrench (force and torque) applied to the object.
Task-specific action spaces are not trivial to implement unless the object is rigidly attached to the end-effector of the robot and its physical properties are known. 
In this work, we aim to learn such an action space and use it for learning multi-robot manipulation tasks that require coordination. 
We propose a method for learning latent central action spaces and learning decentralized policies acting independently in these spaces. 
The previous example is an ideal case where the obtained latent action space has an interpretable physical meaning. 
However, we restrict ourselves to the general case where this shared latent space could also have a semantic uninterpretable meaning, and study its effect on decentralized learning in cooperative multi-robot manipulation tasks.

%% file: L4DC/sections/related.tex
\section{Related Work}
\label{sec:related}

\paragraph{Multi-agent cooperative control} 
MARL methods assign different agents to different parts of the action and state spaces. This reduces the complexity of learning and exploration of the individual components, but the overall problem remains very challenging. 
MARL solutions could be simplified using custom policy parametrizations such as finite state controllers~\citep{bernstein2009policy,amato2010optimizing} or transforming the problem to enable tractable planning and search~\citep{pmlr-v80-dibangoye18a, dibangoye2016optimally}. 
However, decentralized MARL methods fail to achieve a level of coordination, which is needed for physical systems control. 
Hence, several approaches have been proposed to enable a feasible exchange of information during control. 
This could either be achieved through explicit communication channels such as in~\cite{guestrin2002coordinated, sukhbaatar2016learning,singh2018learning,das2019tarmac,pretorius2020learning,niu2021multi}, or via implicit information exchange as part of the policy architecture or the learning algorithm~\citep{gupta2017cooperative,Lee2020Learning,lowe_multi-agent_2017}. 
For instance, multiple methods are based on modeling the other agents' policies~\citep{raileanu2018modeling,liu2020multi,yu2021model}. 
This kind of method is usually based on the centralized training decentralized execution (CTDE) paradigm, where training each policy can benefit from the information that is usually exclusive to the other agents at execution time. 
Others have proposed using CTDE for learning a central dynamics model and use the model for training decentralized policies~\citep{zhang2021centralized,willemsen2021mambpo}. 
Similarly, \citet{lowe_multi-agent_2017} and~\citet{foerster2018counterfactual} propose training decentralized actors using a centralized critic. 
Another common approach is to decompose the value function to the different agents~\citep{sunehag2017value, rashid2018qmix}. 
Beyond CTDE, multiple other solutions have been proposed for alleviating the non-stationarity of MARL tasks. 
For instance, \citet{liu2021collaborative} propose engineering the reward function to punish competitive actions taken by individual agents. 
\citet{gupta2017cooperative} relied on policy parameter sharing across agents, which allows multiple agents to use the same policy network while passing an agent index as part of the observation. 
A more extensive overview on MARL can be found in~\citep{zhang2021multi}.


\paragraph{Latent action representation} 
Different control settings enable different kinds of behavior. For instance, motion control is ideal for reaching a given goal but not perfectly suited for manipulating objects or applying forces. 
Similarly, in RL,  previous work showed that the choice of action space representation could lead to improvements in sample efficiency, energy consumption, robustness~\citep{martin-martin_variable_2019, bogdanovic_learning_2020, aljalbout2021learning, varin_comparison_2019,ulmer2021learning} or learning speed~\citep{peng_learning_2017}.
Types of action representation include torque, joint PD-controller, inverse dynamics, muscle activation or variable impedance control~\citep{peng_learning_2017, varin_comparison_2019, martin-martin_variable_2019, bogdanovic_learning_2020}, but also DMPs can be seen as an action representation~\citep{buchli_learning_2011, schaal_dynamic_2006}.
\citet{bahl2020neural} embed DMPs in the action space by having the policy outputs the DMP parameters.
The studies further show that a relation between the task space and choice of action space is of importance~\citep{martin-martin_variable_2019}.
\citet{ganapathi2022implicit} integrate a differentiable implementation of forward kinematics in neural networks to combine cartesian and joint space control.
More recent publications also show the possibility of learning these action representations from interaction with the environment.
\citet{zhou_plas_2020} and \citet{allshire2021laser} learn a conditional variational autoencoder in order to obtain a latent action space.
Policy search is then performed in this latent action representation.
During inference, the decoder part of the auto-encoder is used to transform the latent action back into the original action space.
\citet{zhou_plas_2020} additionally emphasizes constraining the policy to be within the support of the dataset.
\citet{karamcheti2021lila} use language embeddings to inform the learning of latent action spaces.
\citet{rana2022residual} learn a latent skill-based action space where the skills run at a higher frequency than the policy actions.


%% file: L4DC/sections/method.tex
\begin{figure}
    \centering
    \includegraphics[width=\textwidth]{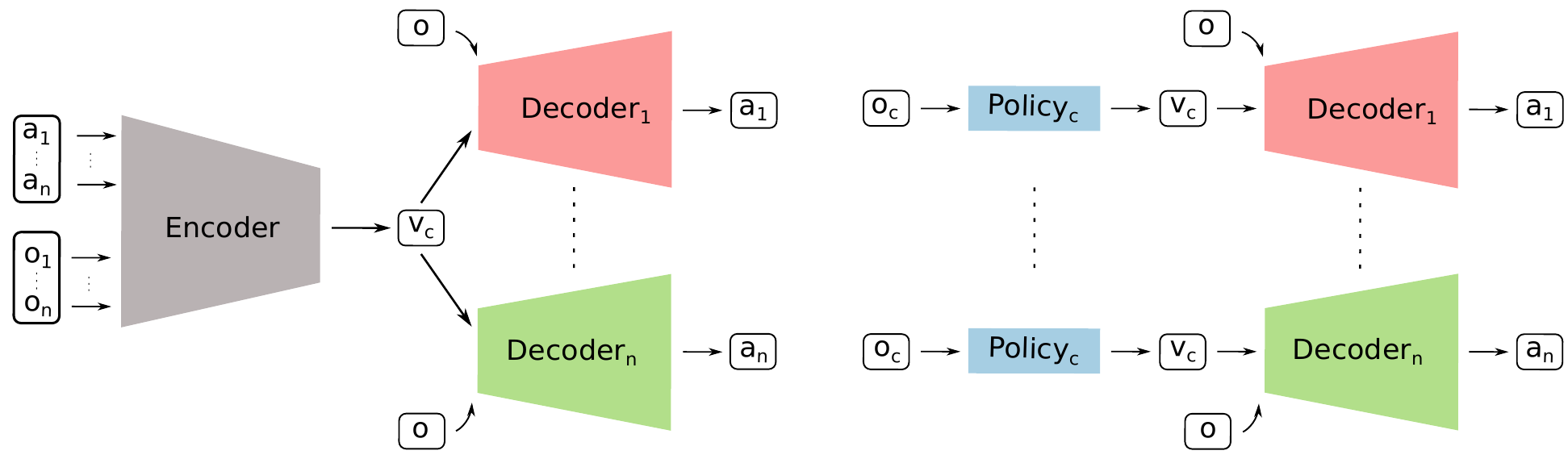}
    \caption{System overview under full agent observability. (Left) we use a conditional autoencoder  for learning the central latent action space. The encoder receives all observations and actions from all agents and produces a latent action $v_c$. This latent action together with the full observation is given to the agent-specific decoders together with the observation. Each decoder outputs an action that is in the original action space of the corresponding agent. (Right) All agents share the same policy acting in the latent action space. The learned decoders map the latent action into the original action space.}
    \label{fig:sysfull}
    \vspace{-1em}
\end{figure}

\section{Method}
\label{sec:method}

We are mainly interested in learning multi-robot manipulation tasks. Typically, such tasks involve multiple robot manipulators (\ie robot arms) that simultaneously interact with an object to achieve a predefined goal. 
Having a single actor control all robots would ideally lead to good coordination, but would suffer in exploration due to the large action and state spaces~\citep{alles2022learning}. Alternatively, control could be split into multiple agents handling one robot each. 
By doing so, we reduce the dimensionality of the individual agents' action and state spaces, hence reducing the sample complexity and exploration requirements. 
However, by decentralizing the process of action generation, coordination between the different agents' policies becomes challenging.
Another main difference to single-agent approaches is the lack of information present to each agent at execution time. 
Namely, each agent can only receive a subset of all the observations of the environment.  
The local agent observations usually correspond to agent-specific observations $\bo_i$ (e.g. proprioceptive measurements in a robotics scenario) and task-related observations $\bo_c$ (e.g. object poses), that are shared across all agents' observations. 

\subsection{Problem Formulation} 
Decentralized cooperative control tasks could be formulated as decentralized partially-observable Markov decision processes (Dec-POMDP). 
A Dec-POMDP is defined by the set $\langle N, \linebreak[1] \mathcal{X}, \linebreak[1] \{\mathcal{U}_i\}_{i \in \{1,\dots,N\}}, \linebreak[1] \mathcal{T}, \linebreak[1] \{r_i\}_{i \in \{1,\dots,N\}}, \linebreak[1] \gamma, \linebreak[1] \{\mathcal{O}_i\}_{i \in \{1,\dots,N\}}, \linebreak[1] \rho \rangle$, where $N$ is number of agents ($N=1$ corresponds to the single-agent problem), $\mathcal{X}$ is the state space shared by all agents, $\mathcal{U}_i$ is the action space for agent $i$, $\mathcal{T}$ represents the environment dynamics, $r_i$ is the reward function for agent $i$, $\gamma$ is the discount factor, $\mathcal{O}_i$ is the observation space of agent $i$, and $\rho$ is the initial state distribution. Since we are interested in cooperative tasks, all agents share the same reward $r_1=r_2=\dots=r_N$. 
Dec-POMDP is a special type of partially observed stochastic games. 
Optimally solving Dec-POMDPs is a challenging combinatorial problem that is NEXP-complete~\citep{bernstein2002complexity}.

\subsection{Central Latent Action Spaces}

Previous approaches relied on the centralized training and decentralized execution paradigm to allow using full observation and action information at least at training time.  
We follow this line of work and propose learning a latent central action space $\mathcal{V}$, which is shared across all agents. 
Controls in this space represent single actions acting on the whole environment and should somehow be translated again into commands to be executed by the individual robots. 
The motivation behind this method is that cooperative tasks usually involve different agents manipulating the same entities to achieve a high-level goal. 
The overall action that is reflected on those entities is a result of all the control commands from all the agents. 
Our approach is illustrated in figure~\ref{fig:sysfull}. We learn a central latent action space using a conditional autoencoder. Given this model, all agents share a single policy to output actions in the latent action space, and translate the given actions to the original action space of each robot based on the learned decoders.

\begin{figure}
    \centering
    \includegraphics[width=\textwidth]{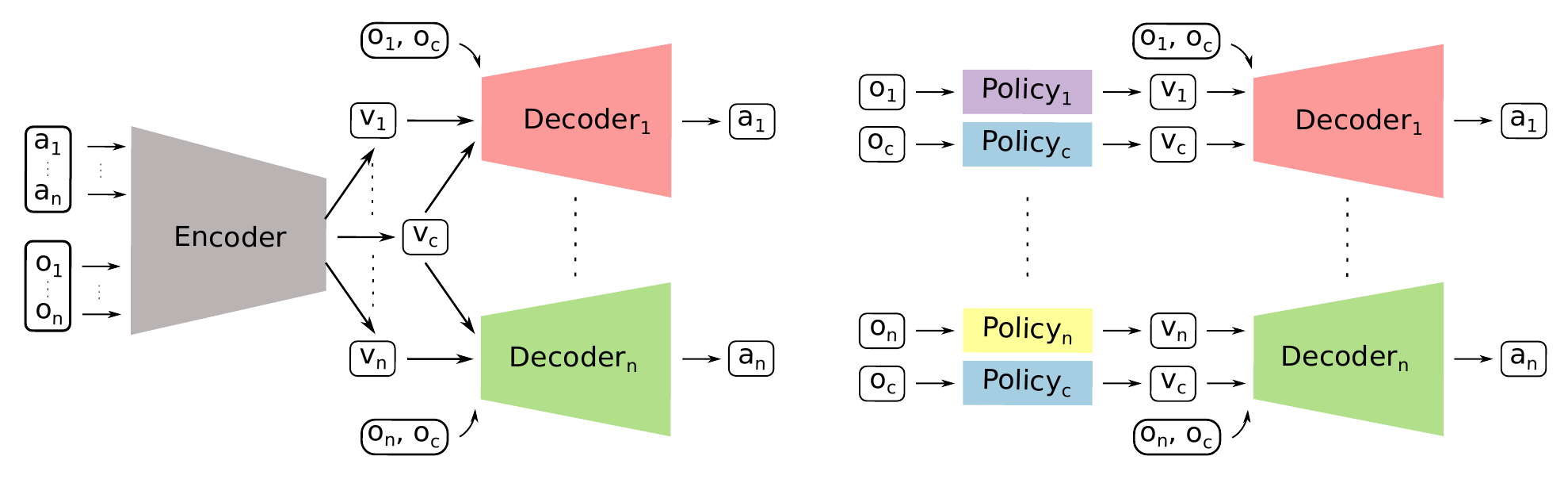}
    \caption{System overview under partial agent observability. (Left) we use a conditional autoencoder  for learning the central latent action space. The encoder receives all observations and actions from all agents and produces a latent action $v$. The latent action contains agent-specific actions $v_i$ as well as a central latent action $v_c$. This latent action together with each agent's observation are given to the agent specific decoders together with the observation. Each decoder outputs an action that is in the original action space of the corresponding agent. (Right) All agents share the same policy acting on the object in the latent action space. They each have a separate policy acting in the latent agent-specific action space. We use the learned decoders to map the latent action into their original action space.}
    \label{fig:syspartial}
    \vspace{-1em}
\end{figure}

To learn a latent central action space, we use Stochastic Gradient Variational Bayes~\citep{kingma2013auto} to overcome the intractable inference distributions involved in learning mappings to this space.
First we look at the case where each agent receives full observations $\bo \in \mathcal{O}_1 \times \mathcal{O}_2 \times \dots \times \mathcal{O}_N$. 
For that, we introduce the graphical models in figure~\ref{fig:sysfull}. 
The generative process of each agent's original action $\bu_i$ is conditioned on the latent central action $\bv$ and the observation $\bo$ (figure~\ref{fig:sysfull}). 
The latter is also used during the inference process, as shown in figure~\ref{fig:sysfull}. 
Additionally---to infer latent actions---actions from all agents $\bu=[\bu_1,\dots,\bu_N]$ are needed.
This is possible since the inference/encoder network will not be used for producing actions in the original space at execution time.
Based on this model, all agents could share a copy of the same policy, which outputs a latent central action $\bv$ based on the full observation $\bo$.  
However, they would each have a different decoder to translate the latent action $\bv$ into their original action space. 
This is illustrated in figure~\ref{fig:sysfull} for a hypothetical environment with two agents. 
The extension to more agents is trivial.
Having a shared policy is feasible in this scenario since the latent action space is supposed to have a lower dimensionality than the aggregated action space of all control agents (\eg robots). 
To illustrate this, we go back to the lifting example. 
Controlling the joint velocity of two robots with six degrees of freedom would result in an action space with a dimension of twelve. Instead, controlling the wrench applied to the object only requires an action space with six dimensions. 
Note that this number does not grow with the number of agents or robots. 
We derive a lower bound to the marginal likelihood: 
\allowdisplaybreaks
\begin{align*}
    \p{\bu}{\bo} &= \int \p[\theta]{\bu}{\bo,\bv}\, \p[\psi]{\bv}{\bo} \, d\bv \\
    \ln \p{\bu}{\bo} &= \ln \int \p[\theta]{\bu}{\bo,\bv}\, \p[\psi]{\bv}{\bo} \, \frac{\q[\phi]{\bv}{\bo, \bu}}{\q[\phi]{\bv}{\bo, \bu}} \, d\bv\\
    &\geq \int \q[\phi]{\bv}{\bo, \bu} \ln (\p[\theta]{\bu}{\bo,\bv} ) \frac{\p[\psi]{\bv}{\bo}}{\q[\phi]{\bv}{\bo, \bu}} d\bv\\
    &=\expc[\q[\phi]{\bv}{\bo, \bu}]{\ln \p[\theta]{\bu}{\bo,\bv}} - \kl{\q[\phi]{\bv}{\bo, \bu}}{\p[\psi]{\bv}{\bo}}
    \numberthis
    \label{eq:elbo_posg2}
    \\
    &= \mathcal{L}(\bu, \theta, \phi, \psi \,|\, \bo), 
    \numberthis
    \label{eq:elbo_posg}
\end{align*}
where $\kl{}{}$ is the Kullback-Leibler divergence, $\q{\bv}{\bo, \bu}$ is the approximate posterior distribution:
\begin{align*}
    \q[\phi]{\bv}{\bo, \bu} &= \gauss{\bv; \mu_v, \sigma_v^2 }\\
    [\mu_v, \sigma_v] &= g_{\phi}(\bo, \bu).
    \numberthis
    \label{eq:gauss}
\end{align*}

Since the generative process of each agent's action is distributed, the likelihood is composed of multiple terms:
\begin{equation}
    \p[\theta]{\bu}{\bo,\bv} = [\p[\theta_1]{\bu_1}{\bo,\bv}, \dots, \p[\theta_N]{\bu_N}{\bo,\bv}],
\end{equation}
Where $\theta_i$ refers to decoder parameters for agent $i$, and $\theta=\{\theta_i\}_{i\in N}$.
Note that the prior is conditioned on the observations. 
It is parameterized by $\psi$ and has a policy-like form $\p[\psi]{\bv}{\bo}$. 
We train it simultaneously to the encoder and decoders using the same loss function from equation~(\ref{eq:elbo_posg}). 

As mentioned, agents in a Dec-POMDP have only access to a subset of the observations. 
However, we notice that in most environments, a certain part of the observations is shared across all agents, and that is usually related to either objects in the scene or any kind of other task-specific observations; but not to the agent's embodiment. 
Even when this condition fails, it could be enforced in the learning process.
For instance, in~\citep{pmlr-v164-liu22a}, the latent space is designed to contain information about the object relevant to the task.

We introduce a new set of graphical models, as seen in figure~\ref{fig:syspartial}. 
In this new model, the latent action space is partitioned into $N+1$ parts. 
The first $N$ correspond to latent actions $\bv_i$, which are specific to each agent. 
The last part $\bv_c$ is central and shared with all agents. 
The generative process of each agent's action (in the original action space) is now conditioned on the agent's observation $\bo_i$, the latent agent-specific action $\bv_i$, and the latent central action $\bv_c$. 
As for inference, the whole latent action variable is conditioned on the full observation $\bo$ and the full action $\bu$. 
As in the previous case, using the full observation and action for inference is possible because the encoder would not be used during control.
Instead, each agent has a set of two policies: one policy producing the latent agent-specific action $\bv_i$ based on $\bo_i$; and another policy that is shared across all agents, and which generate the latent central action based on the shared observation $\bo_c$. 
These two latent actions are then concatenated and decoded into the original action space of the agent. 
 We show the architecture of the policy in figure~\ref{fig:graphical_decpomd}. Note that the policy updates also affect the decoder. The new lower bound is very similar to the one in equation~(\ref{eq:elbo_posg}), with the minor difference of:
\begin{equation}
   \p[\theta]{\bu}{\bo,\bv} = [\p[\theta_1]{\bu_1}{\bo_1,\bo_c,\bv_1, \bv_c}, \dots, \p[\theta_N]{\bu_N}{\bo_N,\bo_c,\bv_N, \bv_c}].
\end{equation}

\subsection{Implementation Details:} The encoders, decoders, prior distributions, and policies involved in this method are implemented as multi-layer neural networks using PyTorch~\citep{NEURIPS2019_9015}. 
All distributions are transformed gaussian distributions using a hyperbolic tangent function ($tanh$).
Actor, critic, and prior networks have two hidden layers. 
Encoders and decoders have three hidden layers.
For training, we use the Adam optimizer~\citep{kingma2014adam}. 
Each policy is optimized using soft-actor-critic (SAC)~\citep{haarnoja2018soft}. 
All modules are trained using randomly sampled data from the replay buffer. 
The latter contains trajectories sampled from the previously described multi-agent policy. 
At the beginning of training, we only update the latent action model using random actions in a warm-up phase that lasts for a hundred thousand steps. 
We found this step to help the training performance and stability.

%% file: L4DC/sections/experiments.tex
\begin{figure}
    \centering
    \includegraphics[width=0.31\textwidth]{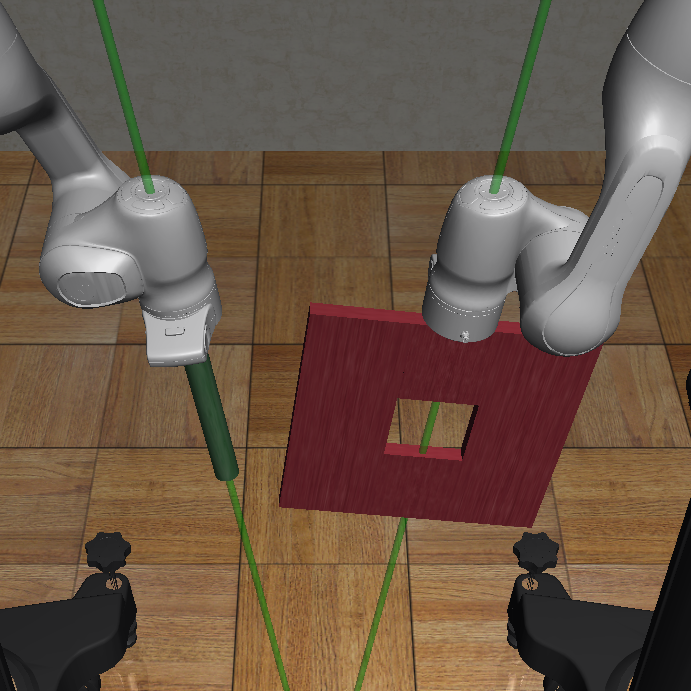} \hfill
    \includegraphics[width=0.31\textwidth]{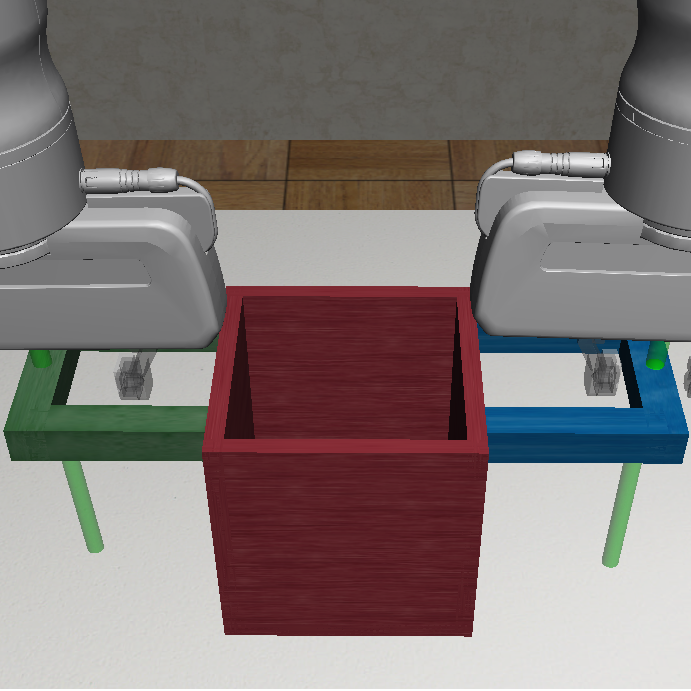}  \hfill
    \includegraphics[width=0.31\textwidth]{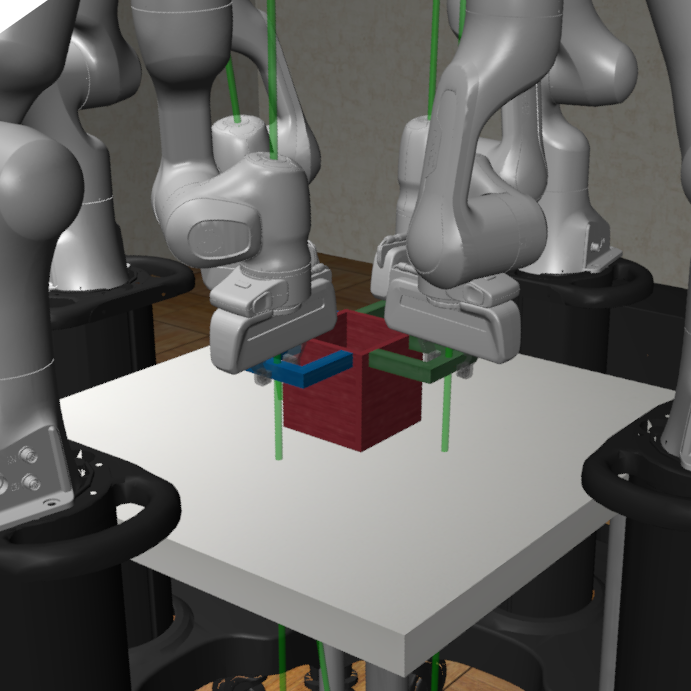}
    \caption{Close-up screenshots from the simulation environments used in our experiments. The environments are provided by robosuite~\citep{robosuite2020}. (left) dual-arm-peg-in-hole environment (middle) dual-arm-lift and (right) four-arm-lift environment with modified gripper structure.}
    \label{fig:environments}
    \vspace{-1em}
\end{figure}

\section{Experiments}
\label{sec:experiments}

We designed our experiments to investigate the following questions:  ($\mathpzc{Q1}$) Can central latent action spaces help coordinate action generation in decentralized cooperative robotic manipulation?
($\mathpzc{Q2}$) Does our method improve sample efficiency with respect to the selected baselines?
($\mathpzc{Q3}$) Can our method reach or exceed the performance of single-agent approaches with full-state information?
($\mathpzc{Q4}$) Is our method scalable to more than two-arm manipulation tasks?
($\mathpzc{Q5}$) How robust is our method to external disturbances?
($\mathpzc{Q6}$) Does our method recover meaningful and task-relevant action representations?

\subsection{Environments} 
\label{sec:setup}

\begin{figure}
    \floatconts{fig:reward_plots_app}{
        \vspace{-1em}
        \caption{
        Episodic rewards in simulated multi-robot manipulation tasks. We compare our method (CLAS) to centralized single-agent and decentralized multi-agent approaches. Our approach outperforms the considered decentralized multi-agent approaches in all environments. It also manages to solve the four-arm-lift task in which all the single-agent and decentralized multi-agent fail.}
        \vspace{-1em}
    }%
    {
        \centering
         \subfigure[dual-arm-peg-in-hole][b]
         {
            \includegraphics[width=0.32\linewidth]{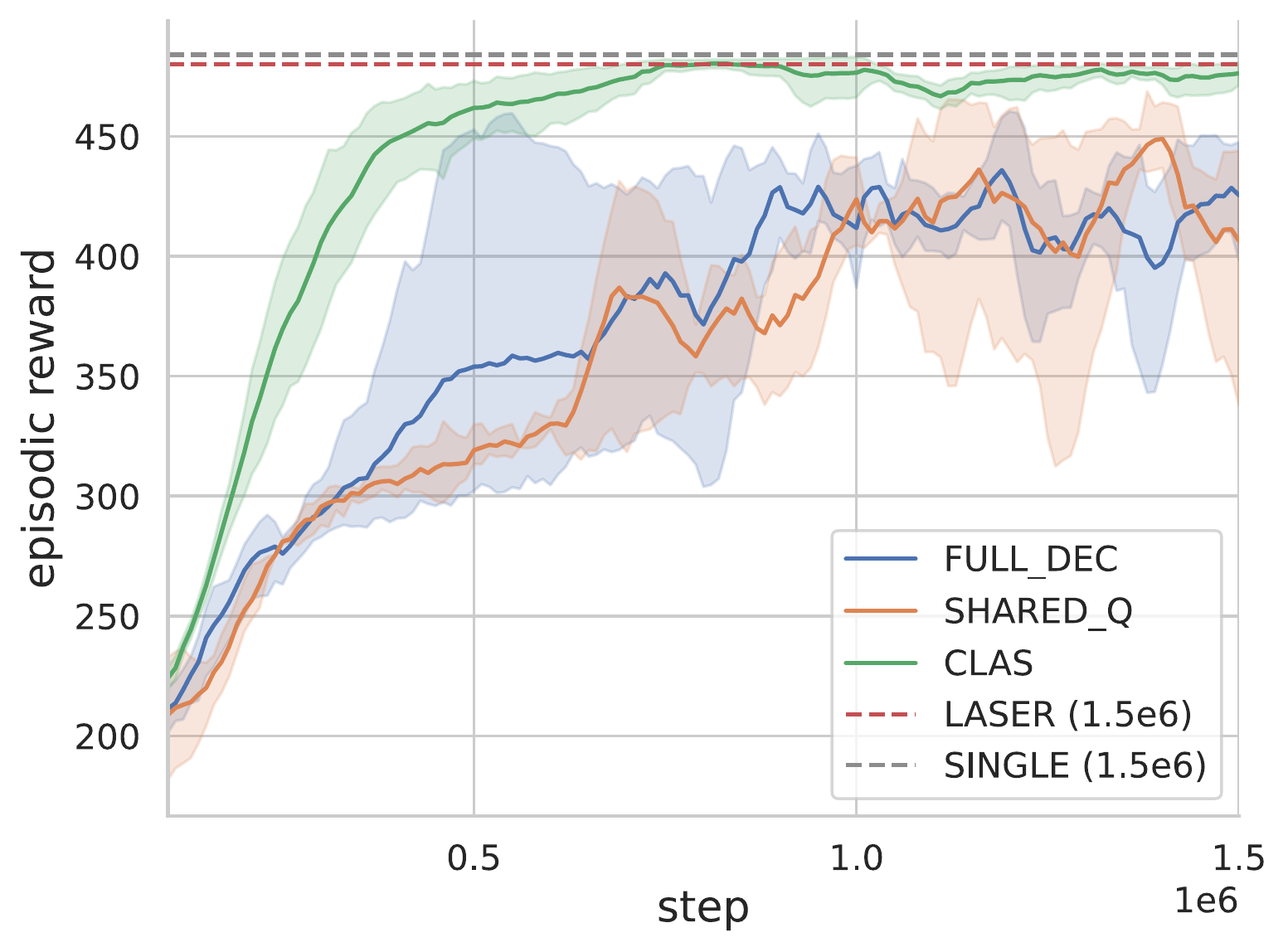}
        }\hfill
        \subfigure[dual-arm-lift][b]
        {
            \includegraphics[width=0.32\linewidth]{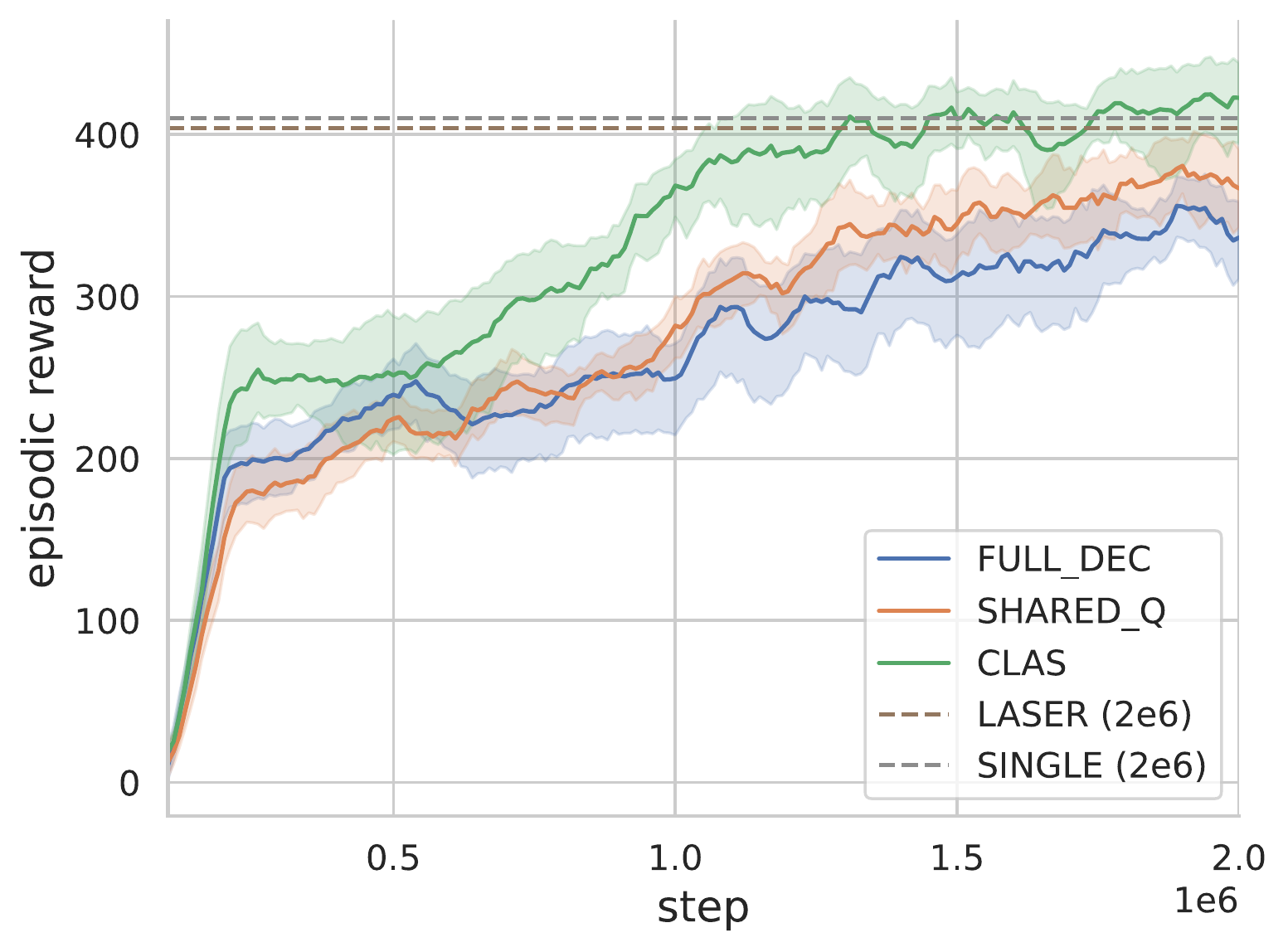}
        }\hfill
        \subfigure[four-arm-lift][b]
        {
            \includegraphics[width=0.32\linewidth]{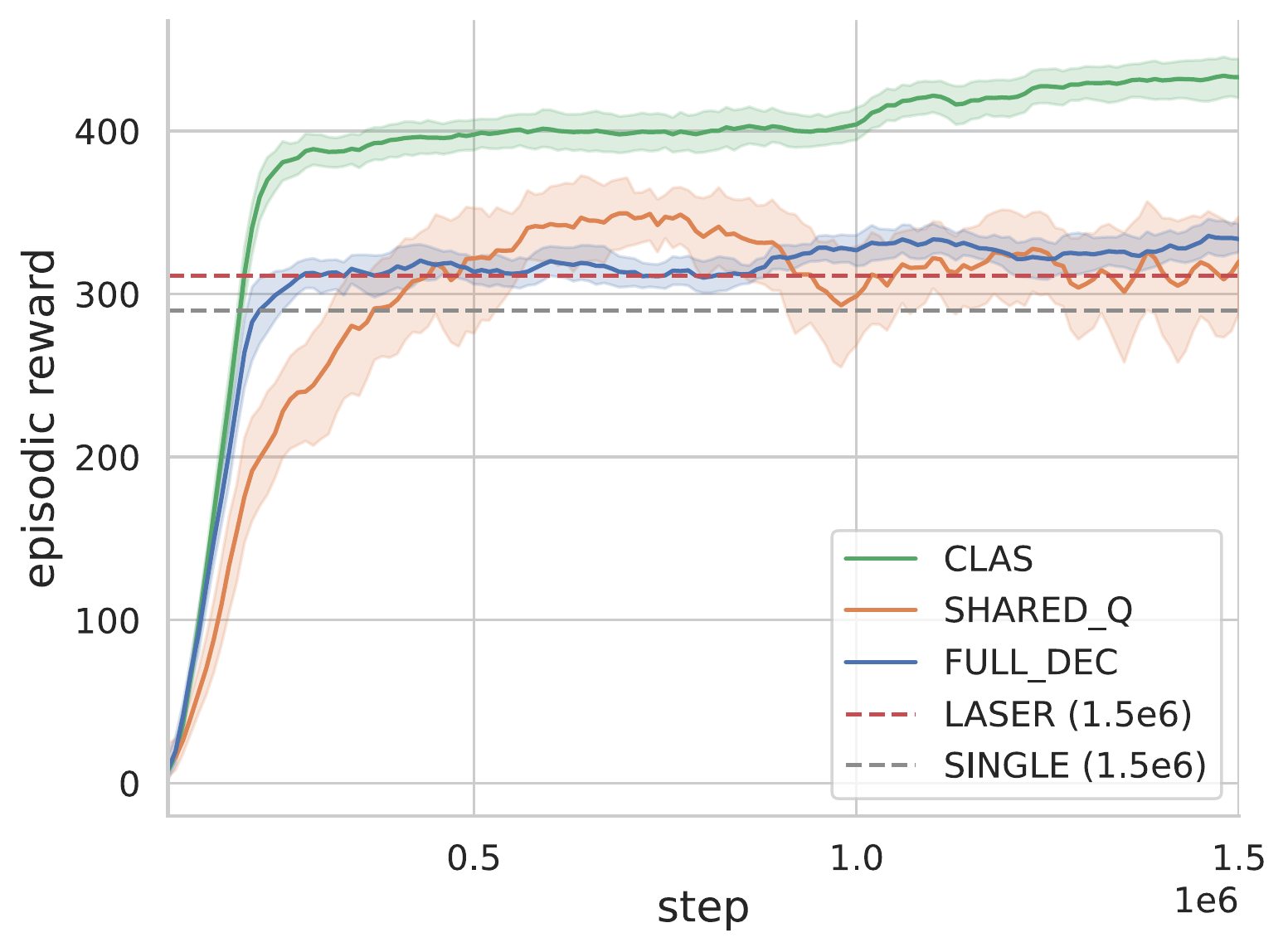}
        }
    }
\end{figure}

We evaluated our method in three simulated environments based on robosuite~\citep{robosuite2020}. The environments are selected/built such that they require cooperation between multiple robot arms. 
Due to the lack of standardized environments that are suitable for our use case (\ie multi-robot manipulation), we use existing environments from public benchmarks when suitable and build alternatives when needed.
Due to the nature of our problem, we select environments that have continuous state and action spaces.
In all of the environments, each agent observations are the corresponding robot's joint position and velocity as well as its end-effector pose.

\paragraph{Dual-arm-peg-in-hole.} In the first environment, two robot arms cooperate on the peg-in-hole task.
A close-up view of the scene and the objects can be seen on the left in figure~\ref{fig:environments}.
We are using the original reward from robosuite, which is composed of a reaching and orientation reward.
The shared observation $o_c$ corresponds to the poses of the peg and hole and the distance between them.


\paragraph{Dual-arm-lift.} For the second environment, we decided to use the dual-arm-lift environment.
In this environment, a rectangular pot with two handles is placed on a surface between two robot arms.
The task for each robot is to reach and grip the handle before cooperatively lifting the pot off the surface.
During initial experiments, we noticed that the provided reward does not promote cooperation and can be easily tricked.
The maximum reward per time step can be reached by controlling a single agent to tilt and lift the pot only slightly off the table.
This is due to the generous maximum tilt angle of $30^{\circ}$ and the successful lift height of $0.10$.
The other major component of the reward measures the ability to reach and grasp the pot handles.
However, we are not interested in assessing the reaching and gripping capabilities but want to rather reward a cooperative lifting behavior.
Therefore we are considering the following modifications to the reward of the environment.
At the start of an episode, we move each robot's end-effector close to its handle and weld the pot handle to the end-effector with a distance constraint in the MuJoCo (\citep{todorov2012mujoco}) simulator.
We chose a distance constraint because it constrains the position but leaves rotational coordinates free.
We remove the gripper fingers to avoid unwanted collisions.
We visualize the resulting starting condition in the middle of figure~\ref{fig:environments}.
We also modify the reward function to enforce success only during high lifts.
Additionally, the maximum tilt angle is reduced such that both robots must cooperate to keep the pot level at all times.
We describe the final reward in equation~(\ref{eq:reward}) in the appendix.
The shared observation $o_c$ corresponds to the pose of the pot.

\paragraph{Four-arm-lift.} The third environment is an extension of the dual-arm-lift environment and uses two additional robot arms to lift the pot (\ie total of four robot arms). Here the pot weight is increased to keep the coordination requirement. We build this environment for the sole purpose of testing scalability to more than two robots/agents.
The pot with four handles and the robot arms' placement can be seen on the right in figure~\ref{fig:environments}.

The changes to the lifting environments were evaluated with manual human control to ensure that tricking the system or solving the task with a single robot arm is not possible. 
Keeping a high reward was only possible when the pot is lifted vertically for a long period of steps.
All environments use a joint velocity controller which receives desired joint velocities from the policy.

\subsection{Baselines:} To validate our method, we compare it to well-established baselines that have been previously applied to continuous control. 
Our experiments include the following baselines:
\begin{itemize}[leftmargin=*, topsep=2pt]
    \itemsep0em 
    \item \texttt{SINGLE}: refers to having a single agent controlling all robots.
    \item \texttt{LASER}: uses a latent action space on top of a single agent controlling all robots. This is based on the work in~\citep{allshire2021laser}. 
    \item \texttt{FULL\_DEC}: refers to having all agents trained with the exact observations and actions they will have access to during execution. The agents are not provided with a communication channel.
    \item \texttt{SHARED\_Q}: similar architecture to \texttt{FULL\_DEC}, but all agents are trained using a central critic. This baseline is based on the work in~\citep{lowe_multi-agent_2017}.
    \item \texttt{CLAS}: refers to our method and abbreviates ``central latent action spaces.''
\end{itemize}
The first two single-agent approaches are included as strong baselines and reference. 
They serve us to better understand the different environments and to elaborately analyze our results. Finally, to make the comparison more reliable, we use SAC for training the different agents in all baselines.

\begin{figure}
    \centering
    \begin{minipage}{0.5\textwidth}
    \centering
    \includegraphics[width=\textwidth]{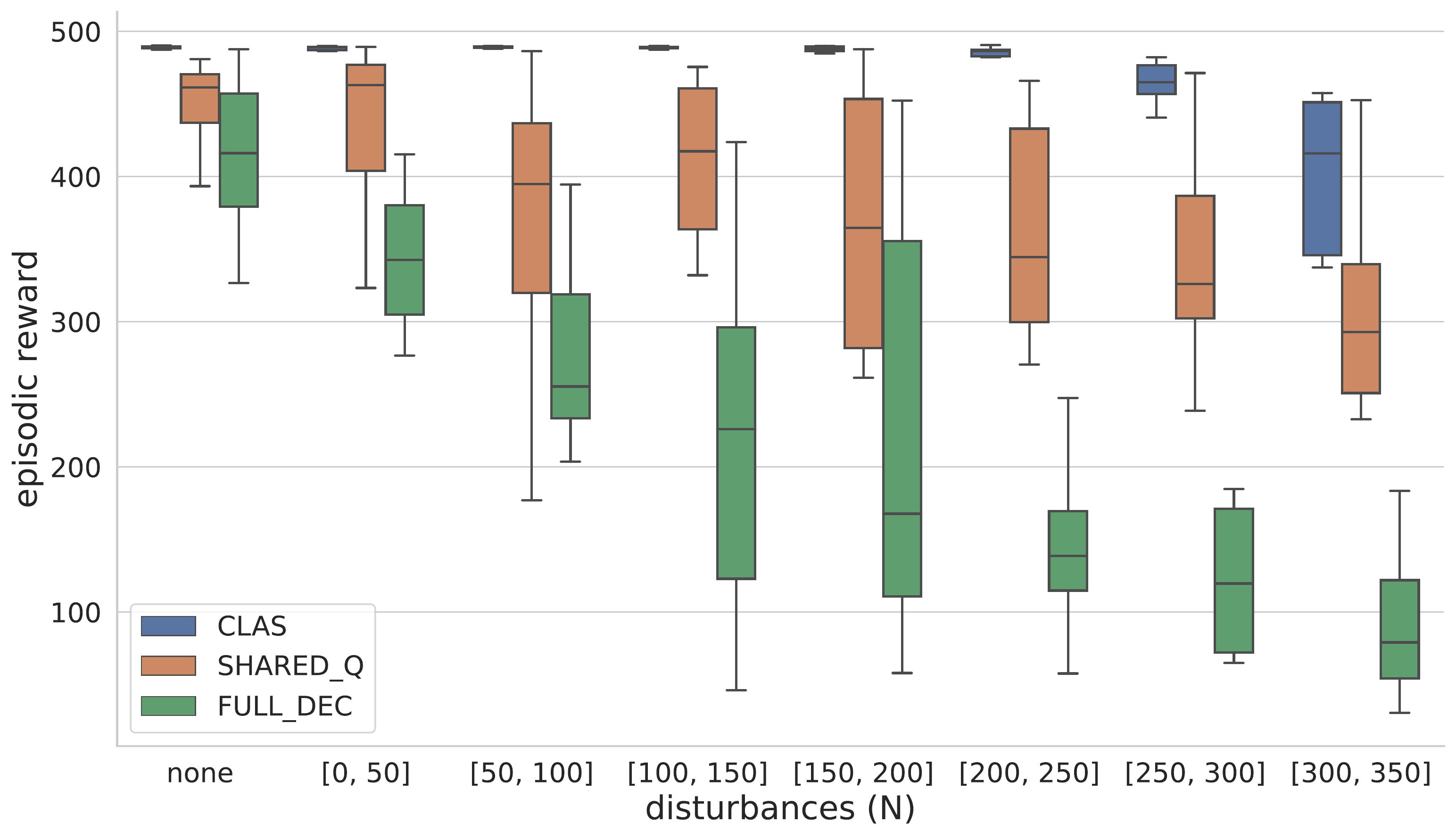}
    \end{minipage}
    \begin{minipage}{0.43\textwidth}
    \centering
    \caption*{Success rates over $10$ evaluation runs under external disturbances.}
     \resizebox{\textwidth}{!}{
    \begin{tabular}{l|ccc}
     & \texttt{FULL\_DEC} & \texttt{SHARED\_Q} & \texttt{CLAS} \\
     \hline \\[-0.8em]
    none & 70\% & \textbf{100\%} & \textbf{100\%} \\
    \textbf{$[0, 50]$} & 30\% & 80\% & \textbf{100\%} \\
    \textbf{$[50, 100]$} & 20\% & 60\% & \textbf{100\%} \\
    \textbf{$[100, 150]$} & 10\% & 70\% & \textbf{100\%} \\
    \textbf{$[150, 200]$} & 20\% & 40\% & \textbf{100\%} \\
    
    \textbf{$[200, 250]$} & 0\% & 40\% & \textbf{100\%} \\
    \textbf{$[250, 300]$} & 0\% & 30\% & \textbf{90\%} \\
    \textbf{$[300, 350]$} & 0\% & 20\% & \textbf{60\%}
    \end{tabular}
    }
    \end{minipage}
        \caption{Effect of applying disturbances (forces) at the center of mass of the pot in the four-arm-lift environment. (left) Episode reward (right) success rate under different ranges of disturbances. Results are based on $10$ evaluation runs. Our method demonstrates robustness against different ranges of disturbances in comparison to the other decentralized baselines, which success rate decreases dramatically as we increase the disturbance.} 
        \vspace{-1em}
    \label{fig:robustness}
\end{figure}

\subsection{Results}

\textbf{Task Performance.} Figure~\ref{fig:reward_plots_app} shows the episodic reward obtained by our method and the baselines on the two considered environments. 
Looking at the single-agent approaches, we observe that both baselines reach high reward areas for the dual-arm tasks. 
However, they both fail to solve the four-arm-lift task.
At the end of training, the best mean episode reward achieved by a single agent is substantially smaller than the maximum possible reward and has a very large variance. 
This illustrates the problem of learning multi-robot manipulation tasks with large action and observation spaces with a single-agent RL approach. 
In contrast to the dual-arm tasks, the four-arm-lift environment features state and action spaces twice the size.

Next, we analyze the results from MARL-based methods. 
\texttt{FULL\_DEC} and \texttt{SHARED\_Q} struggle to keep up with single-agent RL methods. Both methods do not explicitly encourage coordination. Hence, this result might indicate that our environments are well-suited for studying Dec-POMDPs, since they require a certain degree of coordination to be solved. 
The two approaches manage to solve the peg-in-hole task but struggle in the two other environments.
They also lead to very similar results. 
In contrast, our method (\texttt{CLAS}) successfully solves all tasks even under partial observability. 
In the dual-arm-peg-in-hole environment, it reaches a high episode reward after only 250 thousand environment interaction steps, 
while the two other MARL approaches fail to do so in triple the number of steps. 
Furthermore, it achieves a final performance very close to the one achieved by single-agent methods. 
In the dual-arm-lift environment, our approach outperforms both MARL-based baselines. 
Additionally, it surpasses the final performance of the two other MARL approaches after only a half amount of steps. 
More importantly, \texttt{CLAS} slightly outperforms the single-agent methods. 
In the four-arm-lift environment, \texttt{CLAS} is the only studied method that manages to solve the task and achieve a high reward.
Even the single-agent baselines which have access to full state information fail in this task.
This indicates that acting in the latent central action space enables coordinated control even under partial observability and action decentralization.
Finally, we notice that our method leads to significantly lower performance variance, which makes deploying it in real-world scenarios more reliable.

\textbf{Robustness analysis.} We aim to evaluate the coordination capability of our method by quantifying its robustness to external disturbances. We perform this experiment on the four-arm-lift environment and compare the different decentralized baselines to our method. For each method, we pick the model from the training run with the best achieved performance. We then evaluate the corresponding agents in the same environment as before, however, when additionally applying an external force to the pot. The force is applied during the steps in the interval $[10, 100]$ and the values of the force vector are uniformly sampled at each step to be in a certain range. We experimented with multiple ranges. The results can be seen in figure~\ref{fig:robustness}. Under no disturbances ("none"), all methods achieve a high reward and a decent success rate. After applying disturbances in the range $[200-250]$, \texttt{FULL\_DEC} fails in all evaluation runs to solve the task. The success rate of \texttt{SHARED\_Q} goes down to $40\%$, but its reward remains relatively high as the agent manages to lift the pot a bit but not always to the target height. On the other hand, our method \texttt{CLAS} is almost not affected by this level of disturbances. As expected, when increasing the magnitude of the forces, all methods start to fail more often at solving the task, but \texttt{CLAS} appears to remain reasonably robust.

%% file: L4DC/sections/conclusion.tex
\section{Conclusion}
\label{conclusion}

We propose latent central action spaces for decentralized multi-agent control in cooperative tasks. 
The main idea behind our method is to enable coordinated control of multiple robot manipulators based on sharing a latent action space that is agent-agnostic. 
During training time, our approach benefits from central access to all observations and actions and uses this data to train the latent action space model. 
During execution, each agent benefits from the latent central action model to produce control commands that are coordinated with other agents. 
We compare our approach to different baselines and show that latent central action spaces improve the overall performance and efficiency of learning. 
Interestingly, our method solves a task in which centralized baselines struggle. 
Finally, we show that our approach improves robustness against external disturbances.

%% file: L4DC/sections/appendix.tex
\section{Further Details}

\subsection{Models}

We provide further figures illustrating the computational architecture and graphical models related to the different components of the algorithm. Figure~\ref{fig:policygm} shows the grahical models of the policies involved in our method under full and partial observability.

\subsection{Derivations}
\label{appendix:derivations}

Here we go over the derivation of equation~\ref{eq:elbo_posg2} and provide more steps and explanations on how the derivation is performed:

\begin{align*}
    \p{\bu}{\bo} &= \int \p[\theta]{\bu}{\bo,\bv}\, \p[\psi]{\bv}{\bo} \, d\bv \\
    \ln \p{\bu}{\bo} &= \ln \int \p[\theta]{\bu}{\bo,\bv}\, \p[\psi]{\bv}{\bo} d\bv\\
    \ln \p{\bu}{\bo} &= \ln \int \p[\theta]{\bu}{\bo,\bv}\, \p[\psi]{\bv}{\bo} \, \frac{\q[\phi]{\bv}{\bo, \bu}}{\q[\phi]{\bv}{\bo, \bu}} \, d\bv\\
    &\geq \int \q[\phi]{\bv}{\bo, \bu} \ln \Bigl(\p[\theta]{\bu}{\bo,\bv}  \frac{\p[\psi]{\bv}{\bo}}{\q[\phi]{\bv}{\bo, \bu}} \Bigr) d\bv\\
    &=\expc[\q[\phi]{\bv}{\bo, \bu}] {\ln \Bigl(\p[\theta]{\bu}{\bo,\bv}  \frac{\p[\psi]{\bv}{\bo}}{\q[\phi]{\bv}{\bo, \bu}} \Bigr) }\\
    &=\expc[\q[\phi]{\bv}{\bo, \bu}]{\ln \p[\theta]{\bu}{\bo,\bv} - \ln \q[\phi]{\bv}{\bo, \bu} + \ln \p[\psi]{\bv}{\bo} } \\
    &=\expc[\q[\phi]{\bv}{\bo, \bu}]{\ln \p[\theta]{\bu}{\bo,\bv}} - \kl{\q[\phi]{\bv}{\bo, \bu}}{\p[\psi]{\bv}{\bo}}
    \\
    &= \mathcal{L}(\bu, \theta, \phi, \psi \,|\, \bo). 
\end{align*}

The inequality step is based on Jensen's inequality, the pre-last step is due to the product and quotient rules of logarithms, and the last step is based on the definition of the KL divergence. The derivation is in line with the original lower bound derivation for variational autoencoders~\citep{kingma2013auto}.

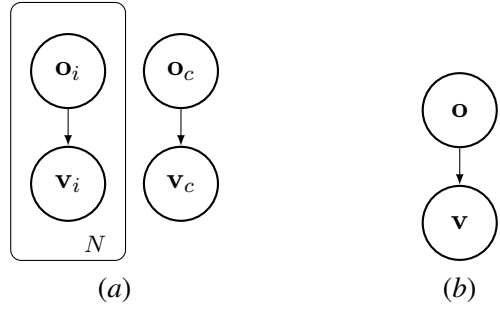
\begin{figure}
    \floatconts{fig:policygm}{
        \caption{
        Graphical models of the policies used by CLAS for the cases of partial (a) and full observability (b).}
    }%
    {
        \subfigure[][b]
        {
            \begin{tikzpicture}[->,auto]
              \tikzstyle{gmnode}=[circle,draw=black,thick,minimum size=28pt,inner sep=1pt]
              \tikzset{>=latex}
              \node[gmnode, draw] (v1) at (-1.5,0) {$\bv_i$};
              \node[gmnode, draw] (vc) at (0,0) {$\bv_c$};
              \node[gmnode, draw] (o1) at (-1.5,1.5) {$\bo_i$};
              \node[gmnode, draw] (oc) at (0,1.5) {$\bo_c$};
              \plate [inner sep=.25cm, yshift=.15cm] {plate1} {(v1) (o1)} {$N$};
              \path
              (o1) edge (v1)
              (oc) edge (vc);
            \end{tikzpicture}
        }
        \hspace{6em}
        \subfigure[][b]
        {
            \begin{tikzpicture}[->,auto]
              \tikzstyle{gmnode}=[circle,draw=black,thick,minimum size=28pt,inner sep=1pt]
              \tikzset{>=latex}
              \node[gmnode, draw] (v) at (0,0) {$\bv$};
              \node[gmnode, draw] (o) at (0,1.5) {$\bo$};
              \path 
              (o) edge (v);
            \end{tikzpicture}
        }
     }
 \end{figure}

 \begin{figure}
    \floatconts{fig:graphical_posg}{
        \caption{
        Graphical models under full access to observations for all agents. (a) action generation, (b) latent action inference. During generation of actions $\bu_i$ each agent $i$ requires input from global observations $\bo$ and central latent actions $\bv$. In order to infer latent actions $\bv$ information from all agents and the global observation is needed.}
    }%
    {
        \subfigure[Graphical model of action generation.][b]
        {
            \hspace{1cm}%
            \scalebox{0.8}{
            \begin{tikzpicture}[->,auto]
              \tikzstyle{gmnode}=[circle,draw=black,thick,minimum size=28pt,inner sep=1pt]
              \tikzset{>=latex}
              \node[gmnode, draw] (u) at (0,-1.5) {$\bu_i$};
              \node[gmnode, draw] (v) at (0,0) {$\bv$};
              \node[gmnode, draw] (o) at (-1.5,0) {$\bo$};
              \plate [inner sep=.25cm,yshift=.15cm] {plate1} {(u)} {$N$};
              \path 
              (v) edge[out=-55, in=55] (u)
              (o) edge[out=-80, in=170] (u);
            \end{tikzpicture}
            }%
            \hspace{1cm}
        }
        \hspace{6em}
        \subfigure[Graphical model of latent action inference][b]
        {
            \hspace{1cm}%
            \scalebox{0.8}{
            \begin{tikzpicture}[->,auto]
              \tikzstyle{gmnode}=[circle,draw=black,thick,minimum size=28pt,inner sep=1pt]
              \tikzset{>=latex}
              \node[gmnode, draw] (u1) at (0,-1.5) {$\bu_i$};
              \node[gmnode, draw] (u2) at (1.5,-1.5) {$\bu_{-i}$};
              \node[gmnode, fill=gray!10, draw] (v) at (0,0) {$\bv$};
              \node[gmnode, draw] (o) at (-1.5,0) {$\bo$};
              \plate [inner sep=.25cm,yshift=.15cm] {plate1} {(u2)} {$N-1$};
              \path 
              (u1) edge[dashed, out=125, in=235] (v)
              (u2) edge[dashed, bend left=25] (v)
              (o) edge[dashed, out=-45, in=215] (v);
            \end{tikzpicture}
            }%
            \hspace{1cm}
        }
     }
 \end{figure}
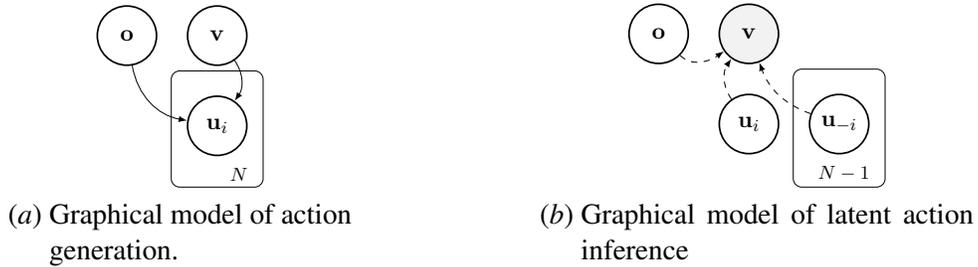
 
 \usetikzlibrary{shapes.geometric, positioning}

  \begin{figure}
    \floatconts{fig:graphical_decpomd}{
        \caption{
        Graphical models under partial agent observability. (a) action generation, (b) latent action inference. During generation of action $\bu_i$ the input observation excludes all the other agents observations $\bo_{-i}$ and latent actions $\bv_{-i}$. Inference is done based on observations and actions from all agents.}
    }%
    {
        \subfigure[Generative action model with latent action space.][b]
        {
            \hspace{1cm}%
            \scalebox{0.8} {
            \begin{tikzpicture}[->,auto]
              \tikzstyle{gmnode}=[circle,draw=black,thick,minimum size=28pt,inner sep=1pt]
              \tikzset{>=latex}
              \node[gmnode, draw] (u) at (-1.5,-1.5) {$\bu_i$};
              \node[gmnode, draw] (v) at (-1.5,0) {$\bv_i$};
              \node[gmnode, draw] (vc) at (0,0) {$\bv_c$};
              \node[gmnode, draw] (o) at (-1.5,1.5) {$\bo_i$};
              \node[gmnode, draw] (oc) at (0,1.5) {$\bo_c$};
              \plate [inner sep=.25cm,yshift=.15cm] {plate1} {(u) (v) (o)} {$N$};
              \path
              (v) edge[bend left] (u)
              (vc) edge[bend left] (u)
              (o) edge[bend right] (u)
              (oc) edge (u.north east);
            \end{tikzpicture}
            }%
            \hspace{1cm}%
        }
        \hspace{6em}
        \subfigure[Graphical model of latent action inference][b]
        {
            \hspace{1cm}%
            \scalebox{0.8} {
            \begin{tikzpicture}[->,auto]
              \tikzstyle{gmnode}=[circle,draw=black,thick,minimum size=28pt,inner sep=1pt]
              \tikzset{>=latex}
              \node[gmnode, draw] (u1) at (-1.5,-1.5) {$\bu_i$};
              \node[gmnode, draw] (u2) at (1.5,-1.5) {$\bu_{-i}$};
              \node[gmnode, fill=gray!10, draw] (v1) at (-1.5,0) {$\bv_i$};
              \node[gmnode, fill=gray!10, draw] (vc) at (0,0) {$\bv_c$};
              \node[gmnode, draw] (o1) at (-1.5,1.5) {$\bo_i$};
              \node[gmnode, draw] (o2) at (1.5,1.5) {$\bo_{-i}$};
              \node[gmnode, draw] (oc) at (0,1.5) {$\bo_c$};
              \plate [inner sep=.25cm, yshift=.15cm] {plate1} {(o2) (u2)} {$N-1$};
              \path
              (u1) edge[dashed, bend left] (v1)
              (u1.north east) edge[dashed] (vc.south west)
              (u2.north west) edge[dashed] (vc.south east)
              (u2.north west) edge[dashed] (v1.south east)
              (o1) edge[dashed] (v1)
              (o2) edge[dashed] (v1)
              (o2) edge[dashed] (vc)
              (oc) edge[dashed] (vc)
              (o1) edge[dashed] (vc)
              (oc) edge[dashed] (v1);
            \end{tikzpicture}
            }%
            \hspace{1cm}%
        }
     }
 \end{figure}
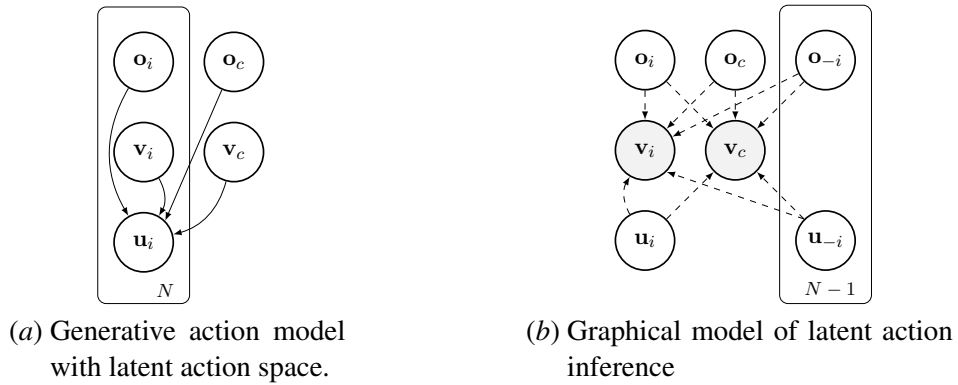

 \section{Experiments}
\label{appendix:exp}

\subsection{Setup}
\label{appendix:setup}
Here we provide further details concerning our setup and experimental design, as to enable easy reproduction of our work.

The environments we used are based on joint velocity control action spaces.
Each agent receives the corresponding robot's proprioceptive measurements, and the shared observation corresponds to object observations.
For evaluation, we run each episode for 500 steps leading to maximal reward of 500. We run all evaluation experiments 10 times with different random seeds.

The reward used in the Lift environment is the following:

\begin{align}\label{eq:reward}
  r_{\text{lift}} &= \max(d - 0.05, 0) \nonumber \\
  r_{dir} &= \left\{
    \begin{array}{lr}
      1, & \text{for } \cos(\alpha) \geq \cos(10^{\circ})\\
      0, & \text{for } \cos(\alpha) < \cos(10^{\circ})
    \end{array}
  \right. \nonumber \\
  r &= \frac{1}{3} \left\{
    \begin{array}{lr}
      3\, r_{\text{dir}}, & \text{for } d > 0.35\\
      10\, r_{\text{dir}} + r_{\text{lift}}, & \text{for } d \leq 0.35,
    \end{array}
  \right.
\end{align}

where $d$ represents the distance between the surface and the pot, $\alpha$ the tilt angle of the pot. 

\subsection{Results under full observability}

Here we study the performance of our method in the case where all agents have access to the full observation. We again compare to the same baselines. Similar to the results in section~\ref{sec:experiments}, our methods outperforms all MARL baselines in terms of final reward and sample efficiency. It also approaches the performance of the centralized single agents, and even outperforms them in four-arm-lift.

\begin{figure}
    \floatconts{fig:resfullobs}{
        \vspace{-1em}
        \caption{
        Results under full agent observability.}
    }%
    {
        \centering
         \subfigure[][b]
         {
            \includegraphics[width=0.47\linewidth]{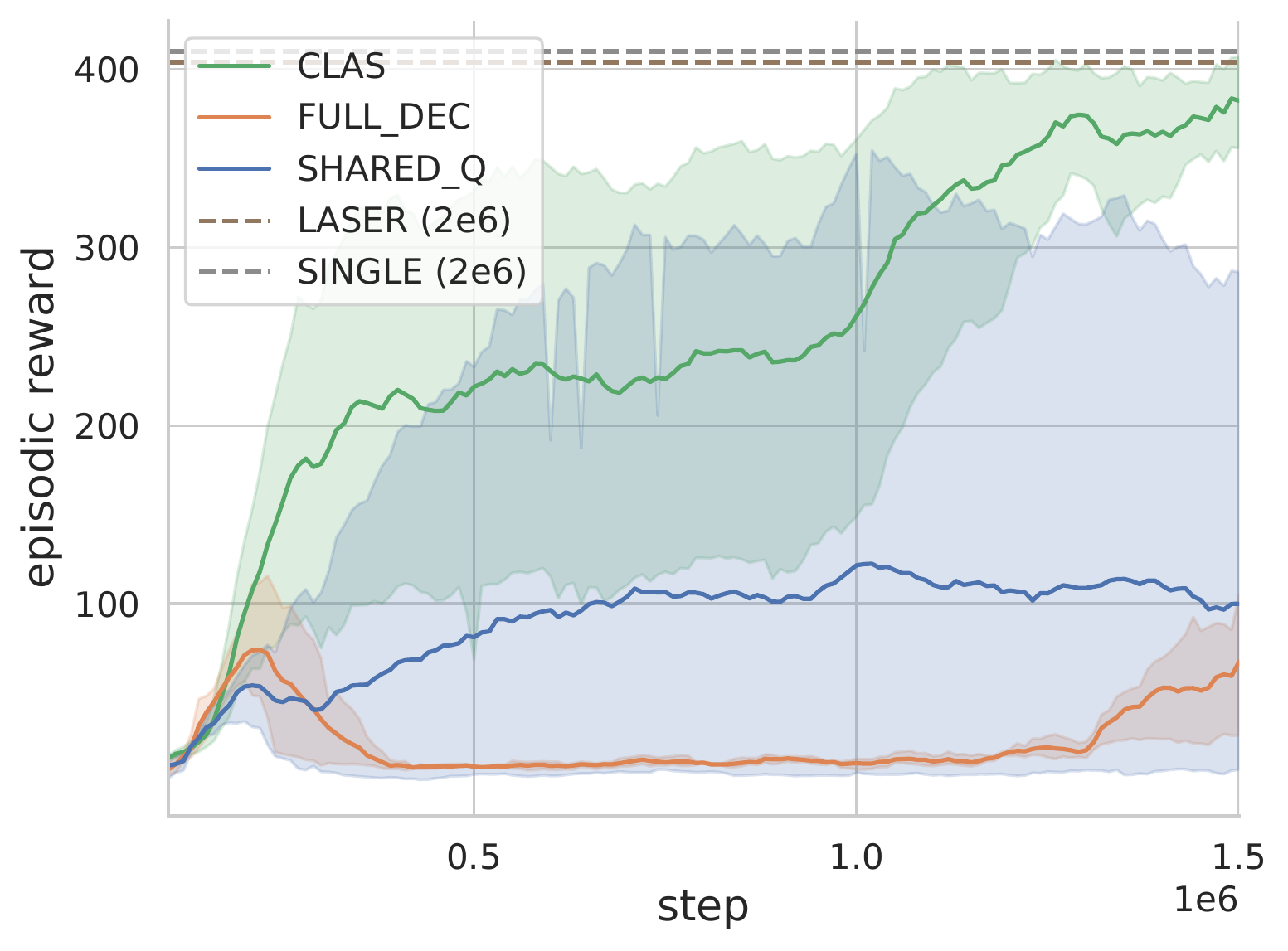}
        }\hfill
        \subfigure[][b]
        {
            \includegraphics[width=0.47\linewidth]{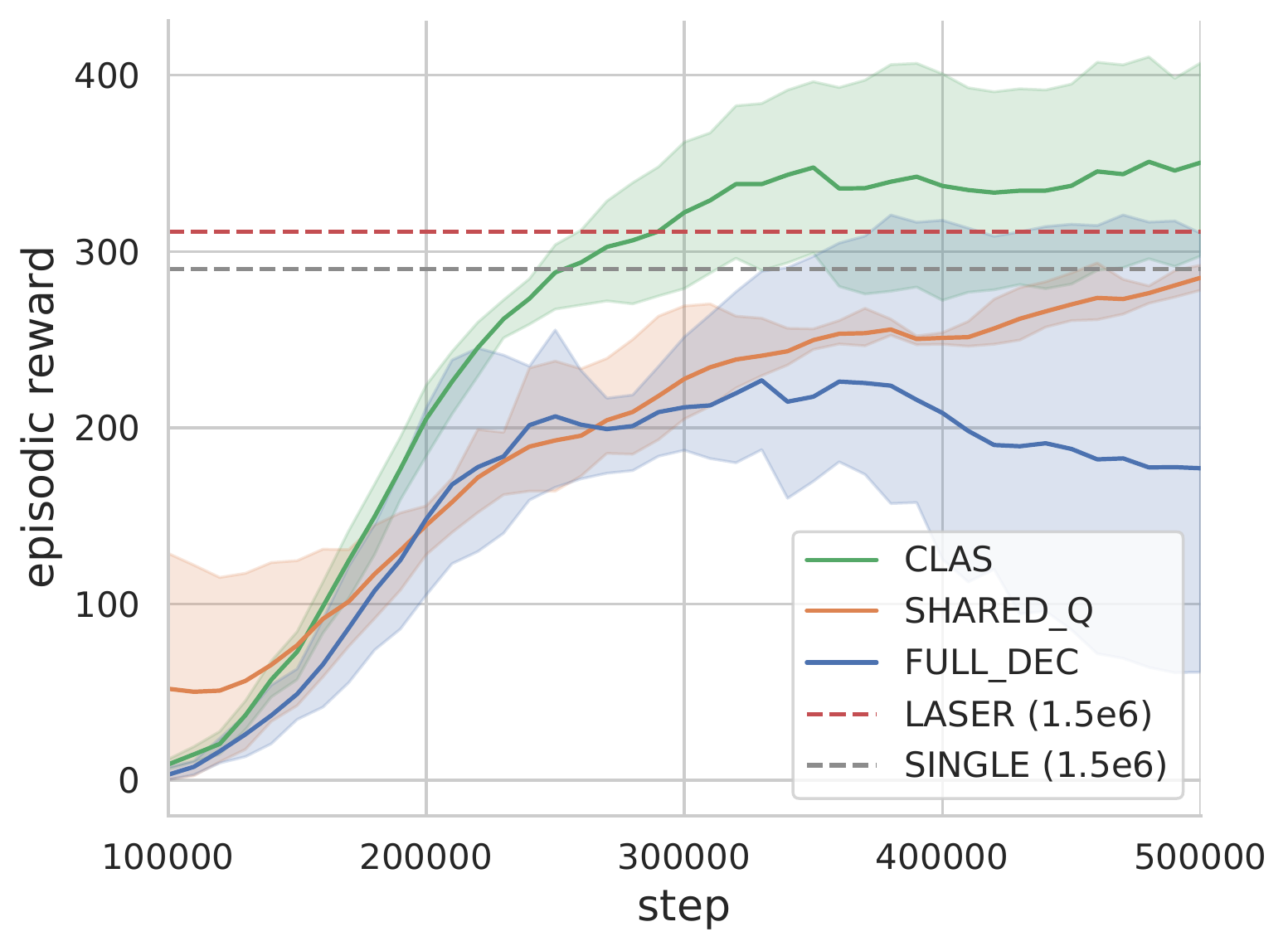}
        }
    }
\end{figure}

\subsection{Ablations}

\begin{figure}
    \centering
    \includegraphics[width=0.7\textwidth]{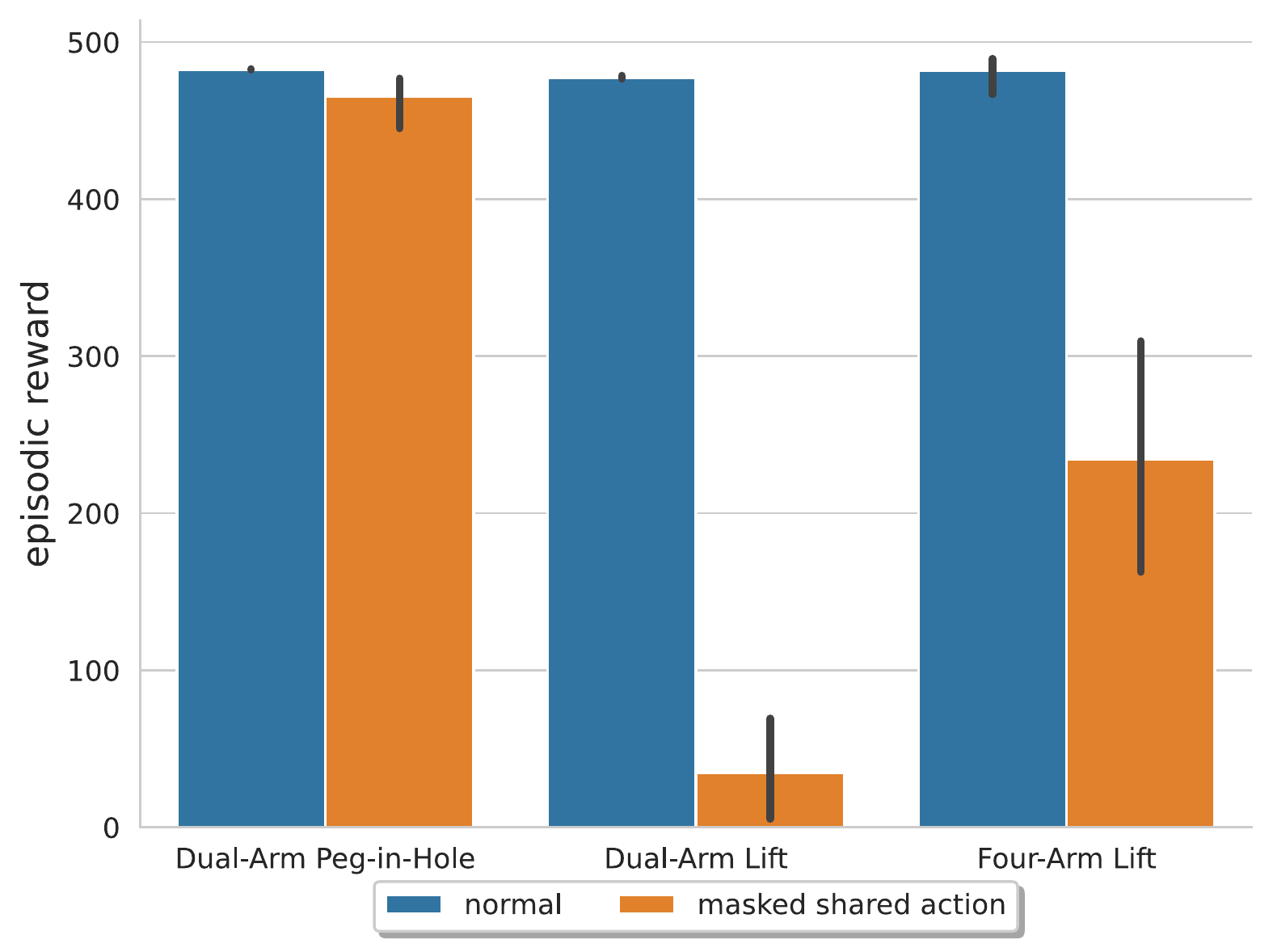}
    \caption{Effect of masking the shared latent action on the achieved total reward.}
    \label{fig:masked_sa}
\end{figure}

In section~\ref{sec:experiments}, we showed that the shared latent actions are active during control. To make sure that the shared latent actions are not ignored during execution we perform the following experiment. We replace the shared latent actions with zeros during inference, and compare the achieved episodic reward to the standard case using our method. The results are in figure~\ref{fig:masked_sa}. For the peg-in-hole environment, the difference in performance is minor. This is mainly due to the fact that this task does not necessarily involve objects that are independent of the robots. Instead the peg and hole are attached to the corresponding robot. The improvement in results shown by our method in figure~\ref{fig:reward_plots_app} is mainly due to the centralized training of the latent action space model. However, for the lifting environments, where a robots-independent object is to be manipulated, masking the shared latent action makes a huge difference. Namely, masking the shared latent actions with zeros leads to very low rewards. These two results indicate that our action space model maps robot actions into actions acting on the objects in some space.


\subsection{Results under Asymmetry of action spaces }

\begin{figure}
    \centering
    \includegraphics[width=0.7\textwidth]{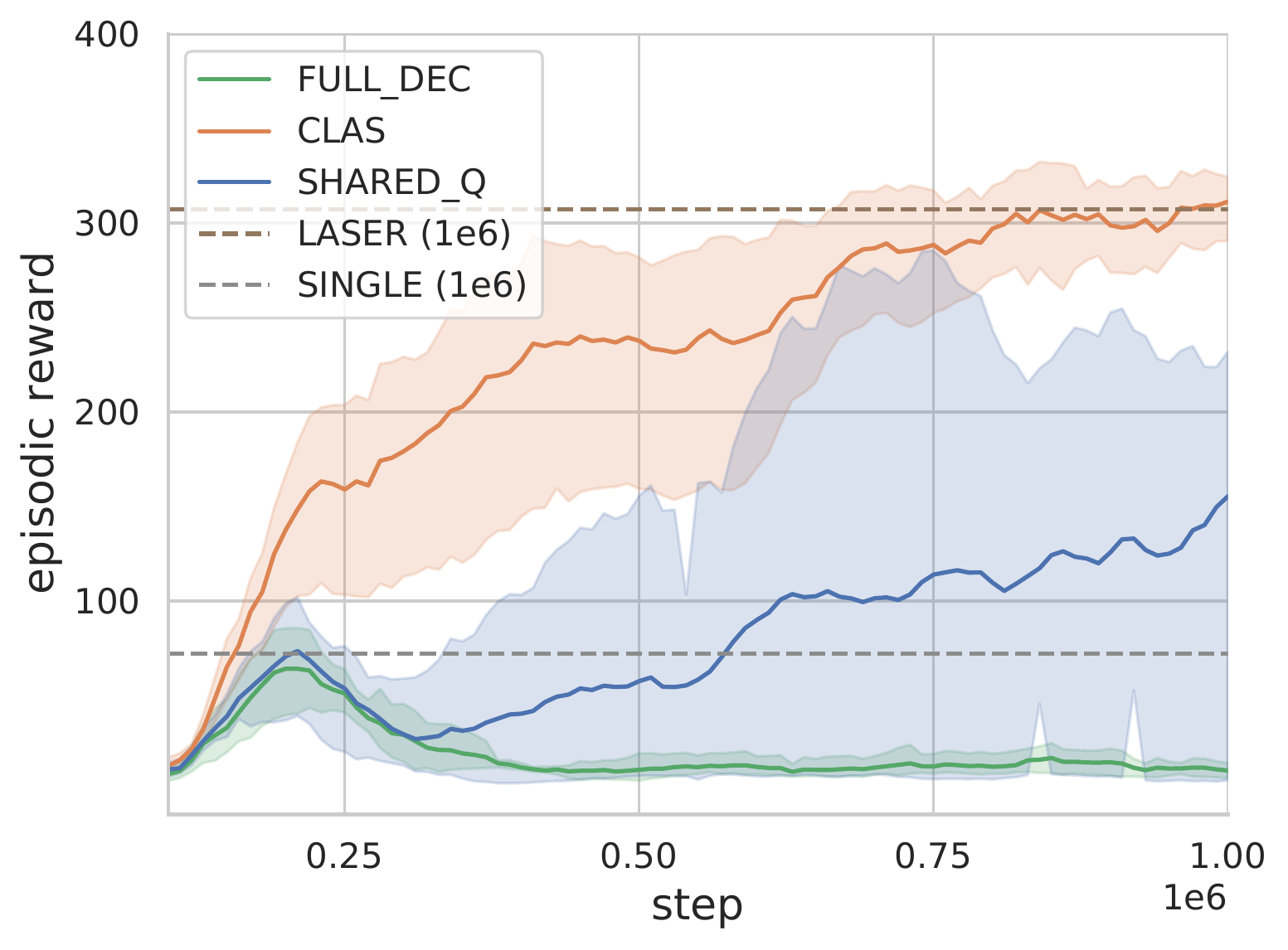}
    \caption{Reward plots in the dual-arm-lift environments when using two different robots with different action spaces.}
    \label{fig:asymetry}
\end{figure}

To check whether our approach is capable of handling asymmetric action spaces and multiple robots, we compare its performance to the baselines again in the dual-arm-lift environments. However, this time we use two different robots in the environment, namely we use a Panda and a Sawyer robot. The panda is equipped with a joint velocity action space and Sawyer with an operational space controller. The results are in figure~\ref{fig:asymetry}. \texttt{CLAS} is the only decentralized method that finds policies capable of lifting the pot, while the other two decentralized baselines as well as \texttt{SINGLE} struggle to do so.

\subsection{Coordination}

\begin{figure}
    \centering
    \includegraphics[width=0.4\textwidth]{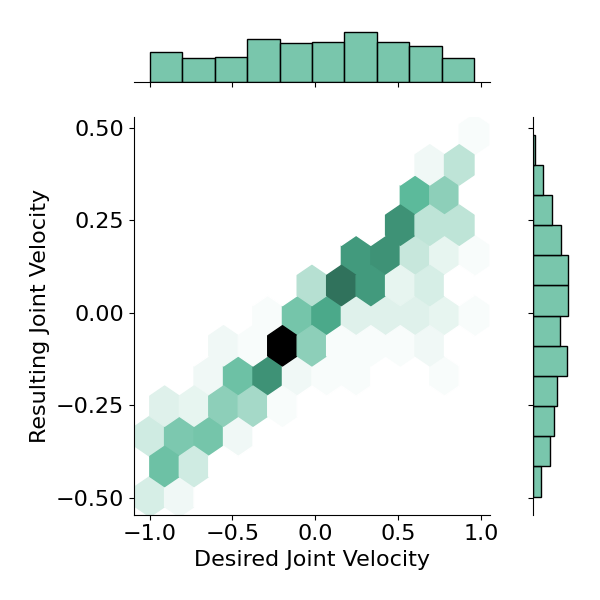}
    \includegraphics[width=0.4\textwidth]{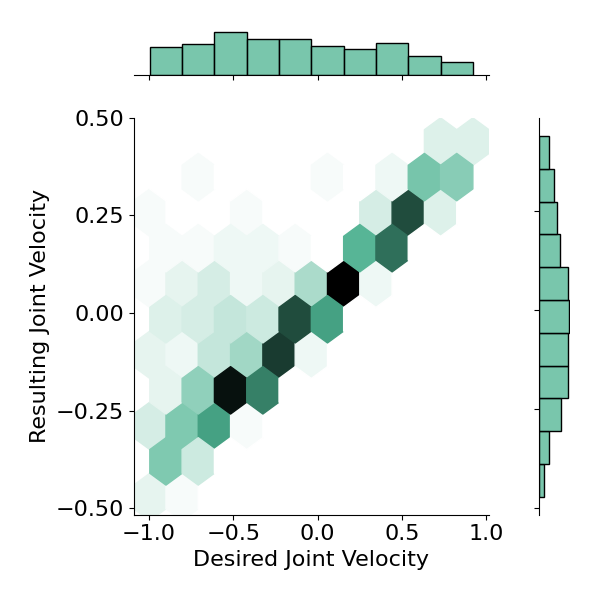} \\
    \includegraphics[width=0.4\textwidth]{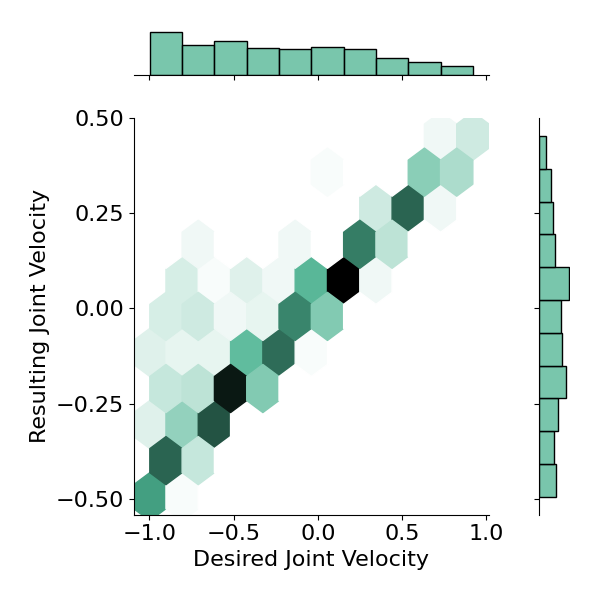}
    \includegraphics[width=0.4\textwidth]{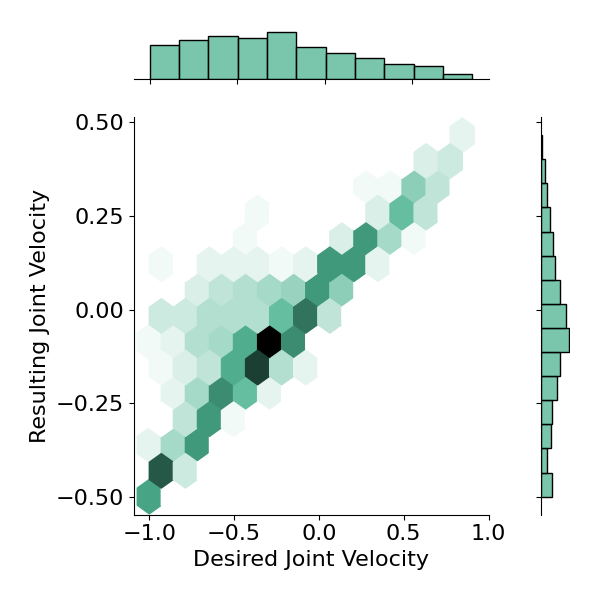} \\
    \includegraphics[width=0.4\textwidth]{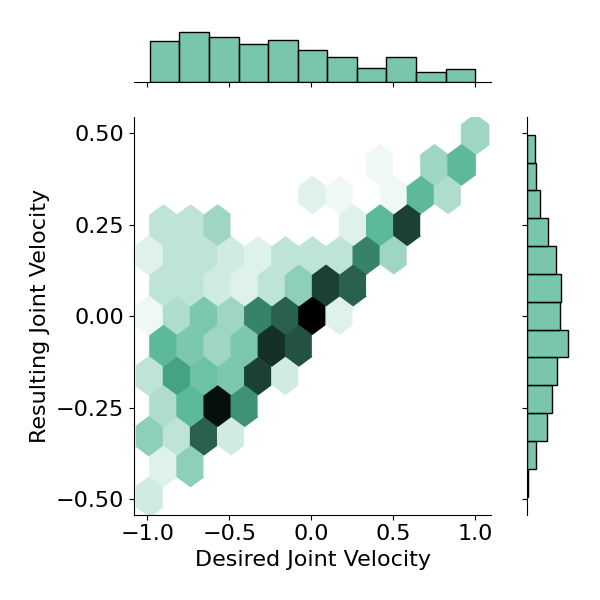}
    \includegraphics[width=0.4\textwidth]{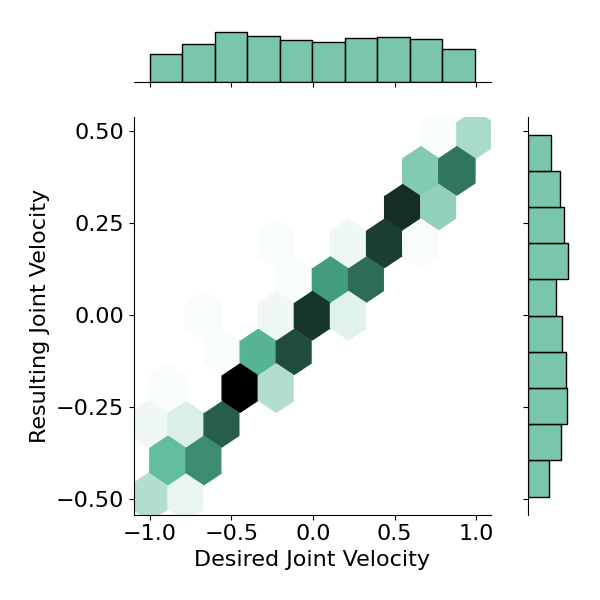} \\
    \includegraphics[width=0.4\textwidth]{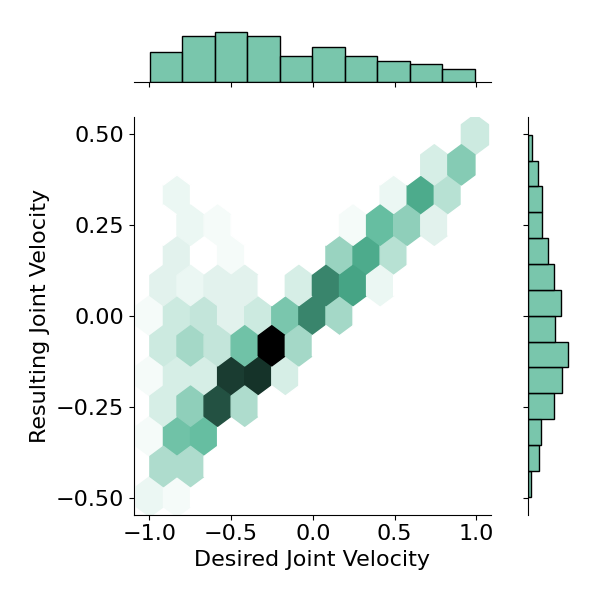}
    \includegraphics[width=0.4\textwidth]{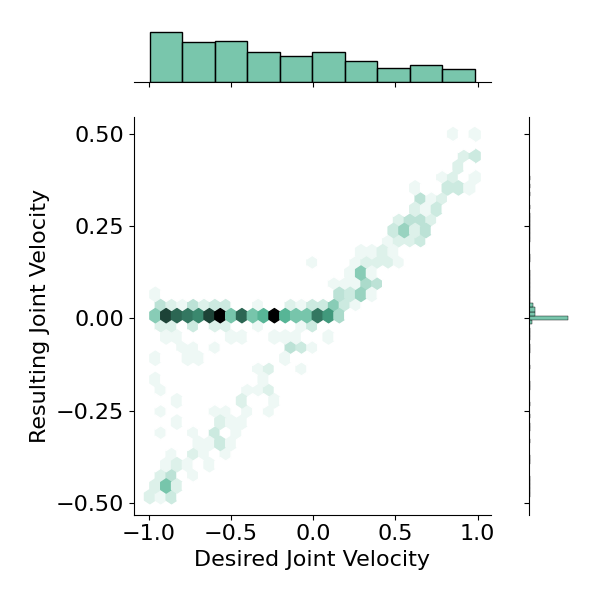} \\
    \caption{Plots of the achieved joint velocity based on the commanded joint velocity for the two agents involved in the Lifting task. Each row indicates a joint [1-4].}
    \label{fig:joints1}
\end{figure}

\begin{figure}
    \centering
    \includegraphics[width=0.4\textwidth]{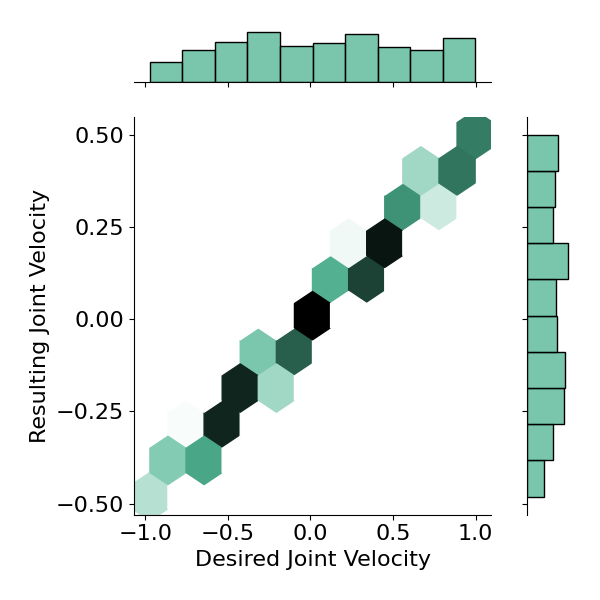}
    \includegraphics[width=0.4\textwidth]{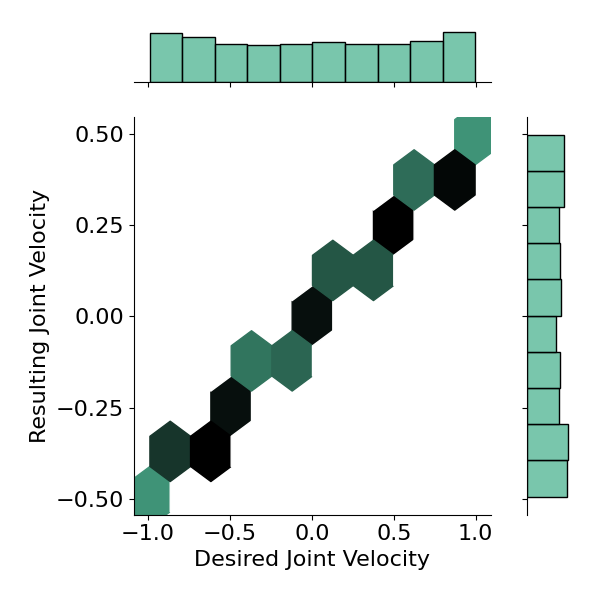} \\
    \includegraphics[width=0.4\textwidth]{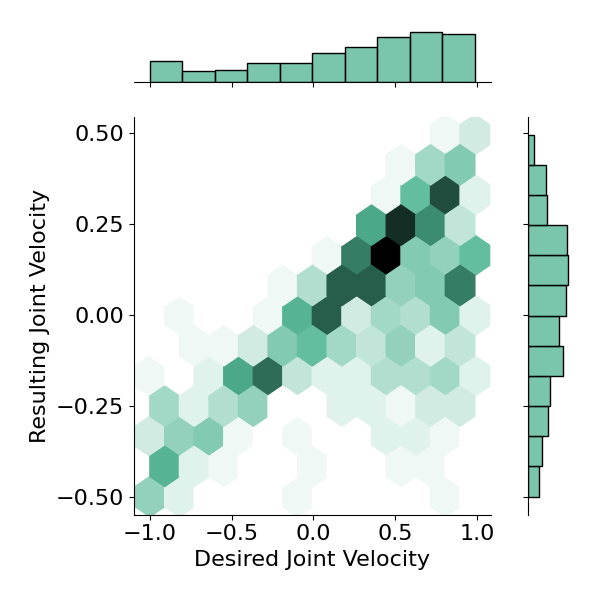}
    \includegraphics[width=0.4\textwidth]{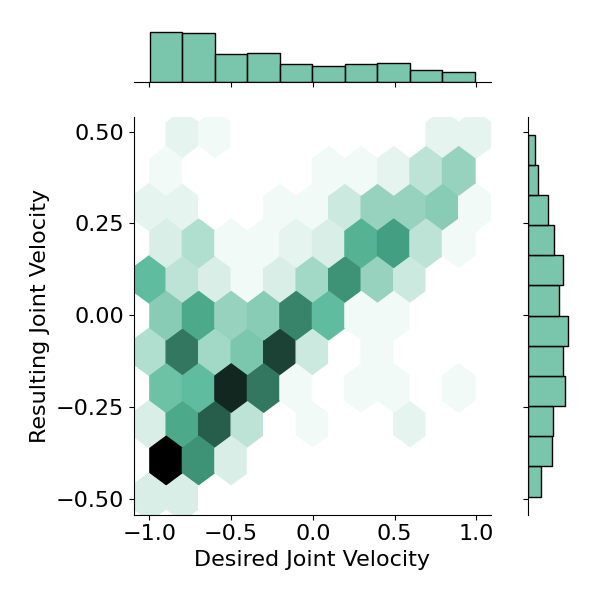} \\
    \includegraphics[width=0.4\textwidth]{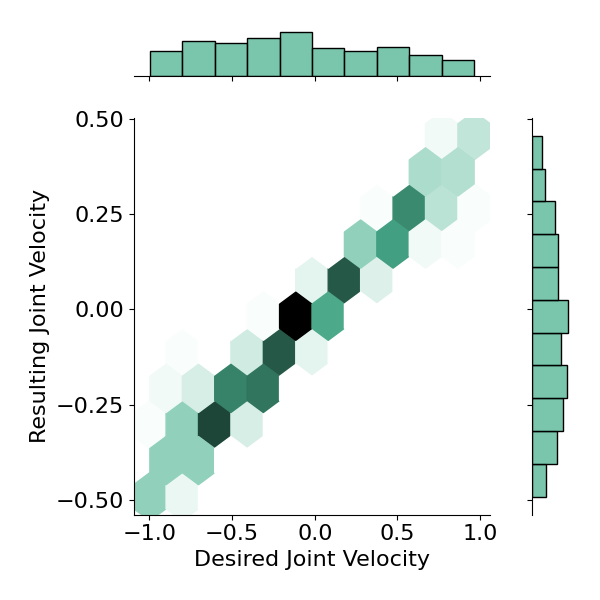}
    \includegraphics[width=0.4\textwidth]{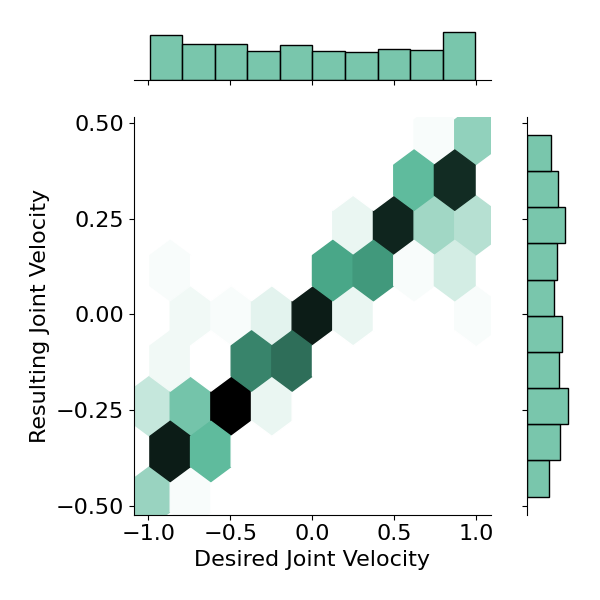}
    \caption{Plots of the achieved joint velocity based on the commanded joint velocity for the two agents involved in the Lifting task. Each row indicates a joint [5-7].}
    \label{fig:joints2}
\end{figure}

To demonstrate the coordination achieved by both agents we plot the desired joint velocity generated by the policy and the achieved joint velocity for both agents. This can be seen in figure~\ref{fig:joints1} and~\ref{fig:joints2}. We notice that the dominant pattern across all plots is the diagonal. This shows that the policy outputs are used by both robots as opposed to having one robot being controlled by a policy, while the other being purely reactive and ignoring its policy outputs. The fourth joint is the only exception, where some policy outputs are ignored (mapped to zeros). However, also in this case, the most values fall on the diagonal.

\subsection{Qualitative Results}

\begin{figure}
    \floatconts{fig:latent_ana}{
        \vspace{-1em}
        \caption{
        Central latent action trajectories for the lifting task. (a)  Trajectories of all latent action dimensions. (b) Correlation between one shared latent action and the z position of the pot. The z-position trajectory of the pot (blue curve) follows the latent action trajectory (red curve).}
    }%
    {
        \centering
         \subfigure[][b]
         {
            \includegraphics[width=0.47\linewidth]{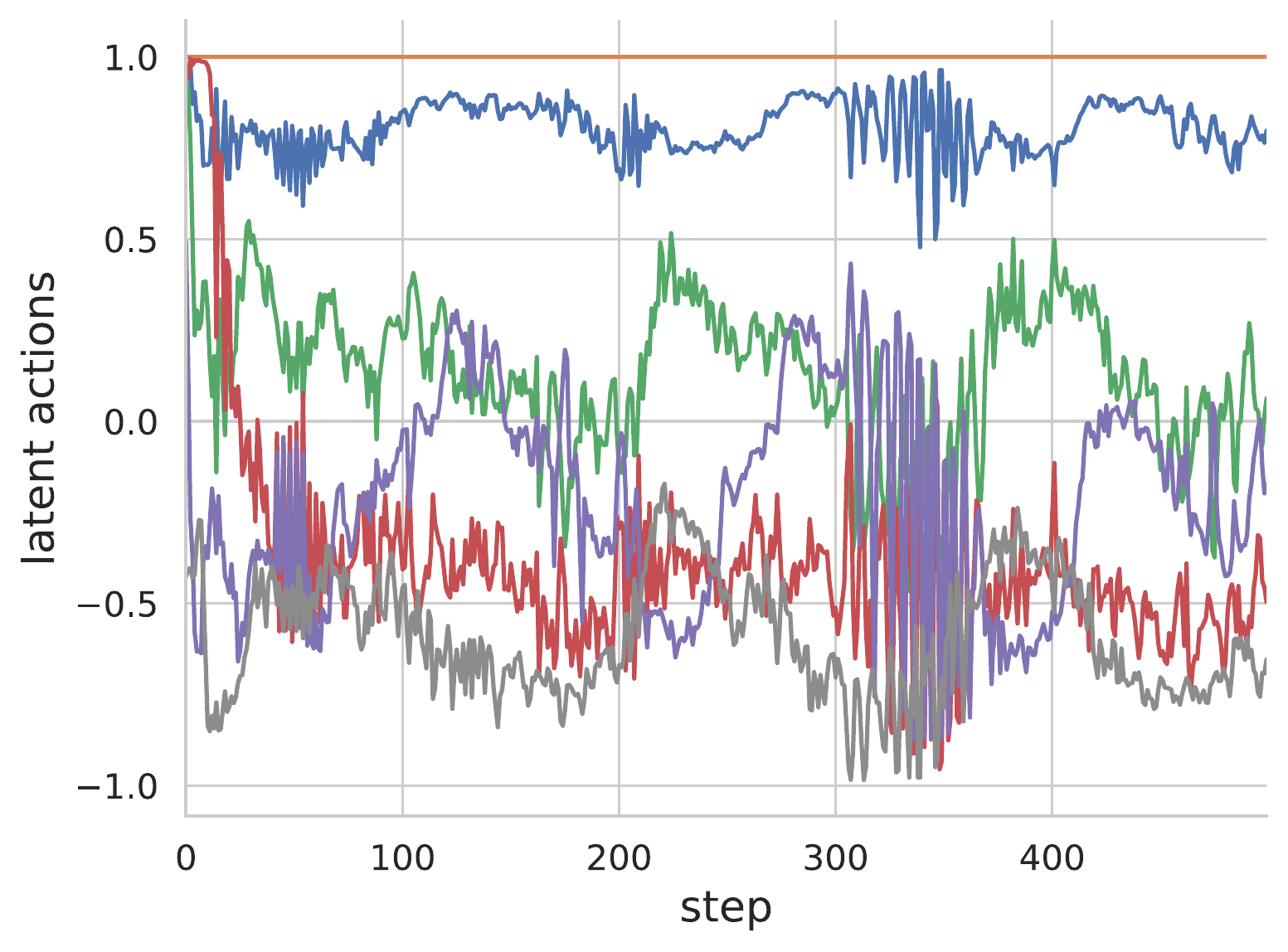}
            \label{fig:action_plotso}
        }\hfill
        \subfigure[][b]
        {
            \includegraphics[width=0.47\linewidth]{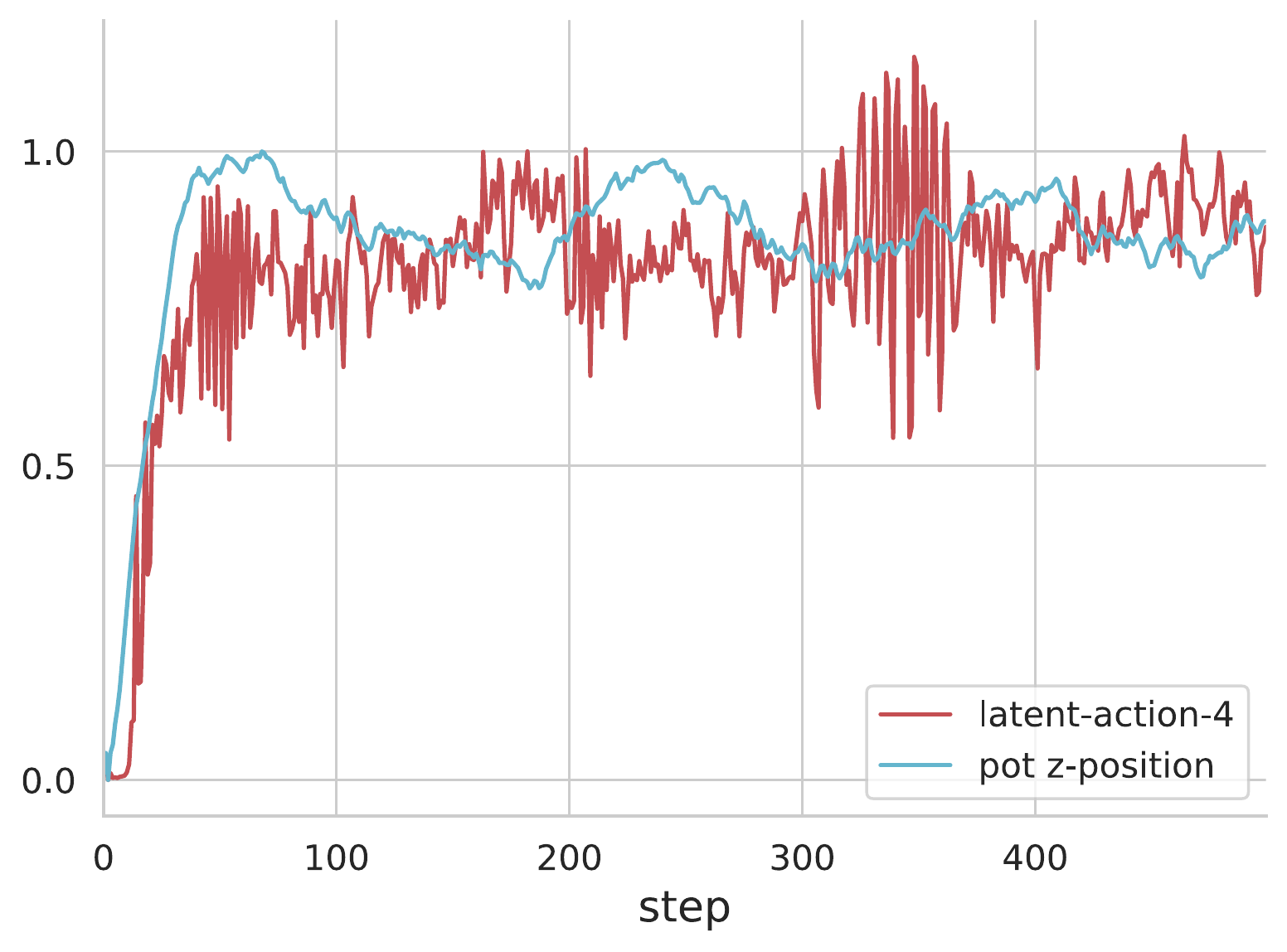}
            \label{fig:z_pot_pos}
        }
    }
\end{figure}

\textbf{Analyzing the central latent action space.} To further validate our method, we examine the shared latent actions produced during evaluation.
Figure~\ref{fig:action_plotso} shows trajectories of the shared latent actions produced by our model for the dual-arm-lift task.
We observe that most shared latent action dimensions are active during control.
One of the latent actions is constant during execution which illustrates that our approach could successfully recover a lower-dimension action space even when configured differently.
Furthermore, we notice that the sequence of actions from the most varying latent action (in red) highly correlates with the z-position trajectory of the pot (figure~\ref{fig:z_pot_pos}). 
The z-position follows the mentioned latent action with a slight time delay. In this case, this latent action represents desired z-positions of the pot needed to lift it.
This is an interesting finding since our approach does not explicitly enforce any physical form or structure on the latent action space. 
The emergence of this property is purely due to the compression capabilities of variational autoencoders. 
Note that the plots in figure~\ref{fig:latent_ana} are qualitative results only meant to illustrate emergent latent actions spaces, and do not mean that our approach is interpretable.

